\documentclass[a4paper, 10pt, conference]{ieeeconf}      % Use this line for a4
\usepackage{FG2025}

\FGfinalcopy % *** Uncomment this line for the final submission

\usepackage{graphicx}
\usepackage{amsmath}
\usepackage{amssymb}
\usepackage{booktabs}

\usepackage{mathtools}
\usepackage[nolist]{acronym} % Basic package for the definition of special characters in sign language
\newcommand{\model}{\epsilon_\theta}
\newcommand{\lsimple}{L_{DM}}
\newcommand{\expec}{\mathbb{E}}
\usepackage{bm}
\newcommand{\ourmethodName}{\textsc{SignDiff}}
\newcommand{\FrameReinforcementNet}{\textsc{FR-Net}}
\newcommand{\FrameReinforcementNetLong}{Frame Reinforcement Network}

\usepackage{caption}
\usepackage[nolist]{acronym}
\begin{acronym}[PHOENIX14\textbf{T} ]

\acrodefplural{rnn}[RNNs]{Recurrent Neural Networks}
\acrodefplural{cnn}[CNNs]{Convolutional Neural Networks}
\acrodefplural{hmm}[HMMs]{Hidden Markov Models}
\acrodefplural{gru}[GRUs]{Gated Recurrent Units}
\acrodefplural{crf}[CRFs]{Conditional Random Fields}
\acrodefplural{gan}[GANs]{Generative Adversarial Networks}
\acrodefplural{gpu}[GPUs]{Graphic Processing Units}

\acrodefplural{mdn}[MDNs]{Mixture Density Networks}

\acro{aslt}[ASLT]{American Sign Language Translation}
\acro{aslp}[ASLP]{American Sign Language Production}
\acro{bsl}[BSL]{British Sign Language}
\acro{bleu}[BLEU]{Bilingual Evaluation Understudy}
\acro{blstm}[BLSTM]{Bidirectional Long Short-Term Memory}
\acro{bslcpt}[BSLCP\textbf{T}]{BSL Corpus \textbf{T}}
\acro{cnn}[CNN]{Convolutional Neural Network}
\acro{crf}[CRF]{Conditional Random Field}
\acro{cslr}[CSLR]{Continuous Sign Language Recognition}
\acro{ctc}[CTC]{Connectionist Temporal Classification}
\acro{c4a}[C4A]{Content4All}
\acro{dl}[DL]{Deep Learning}
\acro{dgs}[DGS]{German Sign Language - Deutsche Gebärdensprache}
\acro{dsgs}[DSGS]{Swiss German Sign Language - Deutschschweizer Geb\"ardensprache}
\acro{dtw}[DTW]{Dynamic Time Warping}
\acro{fc}[FC]{Fully Connected}
\acro{ff}[FF]{Feed Forward}
\acro{gan}[GAN]{Generative Adversarial Network}
\acro{gpu}[GPU]{Graphics Processing Unit}
\acro{gru}[GRU]{Gated Recurrent Unit}
\acro{gslp}[GSLP]{German Sign Language Production}
\acro{gtpt}[G2PT]{Gloss to Pose Transformer}
\acro{hmm}[HMM]{Hidden Markov Model}
\acro{hpe}[HPE]{Hand Pose Enhancer}
\acro{isl}[ISL]{Irish Sign Language}
\acro{lstm}[LSTM]{Long Short-Term Memory}
\acro{mha}[MHA]{Multi-Headed Attention}
\acro{mtc}[MTC]{Monocular Total Capture}
\acro{mse}[MSE]{Mean Squared Error}
\acro{mslp}[MSLP]{Multilingual Sign Language Production}
\acro{mdn}[MDN]{Mixture Density Network}
\acro{mdgs}[mDGS]{meineDGS}
\acro{mdgsv}[mDGS-V]{meineDGS-Variants}
\acro{mdgst}[mDGS\textbf{T}]{meineDGS\textbf{T}}
\acro{mdgsth}[mDGS\textbf{T}-\textbf{H}]{meineDGS\textbf{T}-\textbf{HARD}}
\acro{mdgste}[mDGS\textbf{T}-\textbf{E}]{meineDGS\textbf{T}-\textbf{EASY}}

\acro{nmt}[NMT]{Neural Machine Translation}
\acro{nlp}[NLP]{Natural Language Processing}
\acro{ph12}[PHOENIX12]{RWTH-PHOENIX-Weather-2012}
\acro{ph14}[PHOENIX14]{RWTH-PHOENIX-Weather-2014}
\acro{ph14t}[PHOENIX14\textbf{T}]{RWTH-PHOENIX-Weather-2014\textbf{T}}
\acro{pof}[POF]{Part Orientation Field}
\acro{paf}[PAF]{Part Affinity Field}
\acro{pttt}[P2TT]{Pose to Text Transformer}
\acro{relu}[RELU]{Rectified Linear Units}
\acro{rnn}[RNN]{Recurrent Neural Network}
\acro{rouge}[ROUGE]{Recall-Oriented Understudy for Gisting Evaluation}
\acro{sgd}[SGD]{Stochastic Gradient Descent}
\acro{sla}[SLA]{Sign Language Assessment}
\acro{slr}[SLR]{Sign Language Recognition}
\acro{slt}[SLT]{Sign Language Translation}
\acro{slp}[SLP]{Sign Language Production}
\acro{smt}[SMT]{Statistical Machine Translation}

\acro{ttgt}[T2GT]{Text to Gloss Transformer}
\acro{ttpt}[T2PT]{Text to Pose Transformer}
\acro{ttp}[T2P]{Text to Pose}
\acro{ttgtp}[T2G2P]{Text to Gloss to Pose}
\acro{tts}[TTS]{Text to Speech}
\acro{wer}[WER]{Word Error Rate}

\end{acronym}
\usepackage{dblfloatfix}

\def\B{\fontseries{b}\selectfont}
\usepackage{multirow}

\newcommand{\methodName}{\textsc{SignGAN}}
\usepackage{enumitem,xcolor}
\newlist{coloritemize}{itemize}{1}
\setlist[coloritemize]{label=\textcolor{itemizecolor}{\textbullet},font=\bfseries\color{itemizecolor}}
\usepackage{accents}
\usepackage[accsupp]{axessibility}  % Improves PDF readability for those with disabilities.
\usepackage[colorlinks]{hyperref}
\usepackage[title]{appendix}

\IEEEoverridecommandlockouts                              % This command is only
\usepackage{xspace}
\newcommand{\etal}{\textit{et al.}\xspace}
\newcommand{\ie}{i.e.\xspace}
\newcommand{\eg}{e.g.\xspace}
\newcommand{\etc}{etc\xspace}

\usepackage{fancyhdr}
\def\FGPaperID{317} % *** Enter the FG2025 Paper ID here

\title{\LARGE \bf
SignDiff: Diffusion Model for American Sign Language Production
}

\author{\parbox{16cm}{\centering
    {\large Sen Fang$^1$, Chunyu Sui$^2$$^*$, Yanghao Zhou$^3$, Xuedong Zhang$^4$}\\
    {\large Hongbin Zhong$^5$, Yapeng Tian$^6$, Chen Chen$^7$}\\
    {\small
    $^1$ Rutgers University, $^2$Columbia University, $^3$National University of Singapore, $^4$Victoria University\\
    $^5$Georgia Institute of Technology, $^6$The University of Texas at Dallas, $^7$University of Central Florida\\
    }
    {\normalsize\url{https://signdiff.github.io}}
    \vspace{-24pt}}
}

\begin{document}

\ifFGfinal
\thispagestyle{empty}
\pagestyle{empty}
\else
\author{Anonymous FG2025 submission\\ Paper ID \FGPaperID \\}
\pagestyle{plain}
\fi
%\maketitle

\twocolumn[{
\renewcommand\twocolumn[1][]{#1}%
\maketitle
\begin{center}%
    \centering%
    \includegraphics[width=0.99\textwidth]{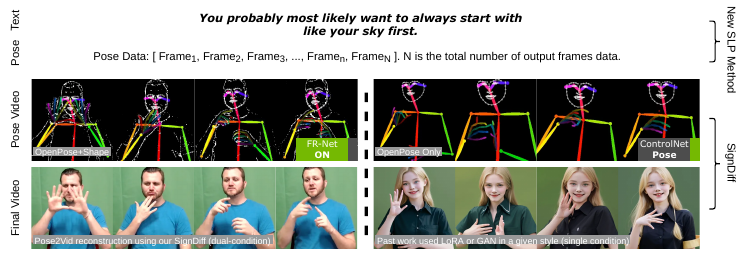}%
    \captionof{figure}{\textbf{Our Sign Language Production:} First, input the English text you want to translate, and our new Text2Pose method will translate the text into continuous skeletal pose sequences. Then we plot the pose data of each frame into appropriate images, and users can further generate the final video using our new pose-conditioned human synthesis model (SignDiff).}%
    \label{fig:intro}%
\end{center}%
}]

%\renewcommand\footnoterule{\rule{0.2\textwidth}{0.2pt}}
%\renewcommand\thefootnote{}
%\footnotetext{$^*$ Collaborator Author.}

% 在导言区添加
\newcommand{\nofootnotenumber}[1]{%
  \begingroup
  \renewcommand{\thefootnote}{}\footnote{#1}%
  \addtocounter{footnote}{-1}%
  \endgroup
}

\renewcommand\footnoterule{\rule{0.2\textwidth}{0.2pt}}
% 使用时
\nofootnotenumber{$^*$ Collaborator Author.}

\thispagestyle{fancy}

%%%%%%%%%%%%%%%%%%%%%%%%%%%%%%%%%%%%%%%%%%%%%%%%%%%%%%%%%%%%%%%%%%%%%%%%%%%%%%%%
\begin{abstract}

%The field of \acf{slp} lacked a large-scale, pre-trained model based on deep learning for continuous American Sign Language (ASL) production in the past decade. This limitation hampers communication for all individuals with disabilities relying on ASL. To address this issue, 
%we undertook the secondary development and utilization of How2Sign, one of the largest publicly available ASL datasets. Despite its significance, prior researchers in the field of sign language have not effectively employed this corpus due to the intricacies involved in \acf{aslp}. 
%To conduct large-scale \ac{aslp}, 
In this paper, we propose 
%based on the latest work in related fields, which is 
a dual-condition diffusion pre-training model named \ourmethodName{} that can generate human sign language speakers from a skeleton pose. \ourmethodName{} has a novel Frame Reinforcement Network called \FrameReinforcementNet{}, similar to dense human pose estimation work, which enhances the correspondence between text lexical symbols and sign language dense pose frames, reduces the occurrence of multiple fingers in the diffusion model.
In addition, we propose a new method for \ac{aslp}, which can generate ASL skeletal pose videos from text input, integrating two new improved modules and a new loss function to improve the accuracy and quality of sign language skeletal posture and enhance the ability of the model to train on large-scale data.
We propose a simple baseline for ASL production and report the scores of 17.19 and 12.85 on BLEU-4 on the How2Sign dev/test sets. We evaluated our model on the previous mainstream dataset PHOENIX14T, and our method achieved the SOTA results. In addition, our image quality far exceeds all previous results by 10 percentage points in terms of SSIM.
%Finally, we conducted ablation studies and qualitative evaluations for discussion.

\end{abstract}

%%%%%%%%%%%%%%%%%%%%%%%%%%%%%%%%%%%%%%%%%%%%%%%%%%%%%%%%%%%%%%%%%%%%%%%%%%%%%%%%
\section{INTRODUCTION}\label{sec:intro}

% Sign Intro
Sign language is a complex language with various challenges for learners in both learning and expression \cite{stokoe1980sign}. In sign language processing (SLP), as shown in Fig. \ref{fig:intro}, textual vocabulary is transformed into skeleton sequences \cite{saunders2021continuous,saunders2020progressive,zelinka2020neural,huang2021towards,brentari2018production} and modeled into 3D human body videos. Some studies also learn gloss, which is a type of sign language annotation that represents text. Previous \ac{slp} approaches combined techniques from neural machine translation (NMT), generative adversarial networks (GANs), pose extraction, and image transformation \cite{stoll2018sign,zelinka2020neural,saunders2020adversarial}. However, data processing in sign language is time-consuming and complex due to the lack of standardization and the need for multiple rounds of dataset processing \cite{7298594,cao2018openpose,guler2018densepose}. How2Sign \cite{duarte2021how2sign} is a recently released multimodal and multi-view continuous American Sign Language (ASL) dataset in 2021, whereas previous sign language production studies mainly utilized the RWTH-PHOENIX-Weather-2014T (PHOENIX14T) German sign language dataset \cite{forster2012rwth}. 
Due to the complexity of data processing, fewer researchers have yet explored ASL production using deep learning techniques on this dataset. 
Compared with PHOENIX14T, the duration and vocabulary of How2Sign are approximately 7.53 and 5.74 times larger than the commonly used PHOENIX14T, respectively.

% Proposed method
In this paper, we propose a large-scale \acf{aslp} method (text2pose) and a diffusion model \cite{Rombach_2022_CVPR} for pose-conditioned human body synthesis (pose2video). 
Firstly, we discard the intermediate conversion stage from text to gloss and directly perform a more efficient translation from spoken language to pose (text2pose), the method will be abbreviated as ``Fast-SLP'' in the following.
Then, we propose a \ourmethodName{} model to learn the correspondence between sign language keypoints and real images representations, enabling high-quality rendering of the generated long sequential videos (pose2video).

%former & signdiff part
% For Fast-SLP Transformers, to accommodate the larger and semantically richer dataset How2Sign. We have developed adaptable modules that optimize mask handling, batch computation, memory utilization, and tensor operations. Additionally, we have designed a novel loss function specifically tailored to the model, which enhances the learning process for longer videos and enables the production of sign language demonstrations of extended durations. Previously, the generated sign language videos were used by our method, called \ourmethodName{}, to create higher-resolution and visually appealing representations of real human signers. \ourmethodName{} is the first sign language pre-training diffusion model obtained by fine-tuning the standardized SD1.5 base model using processed How2Sign video data. The previous GAN-based approach \cite{stoll2020text2sign,cui2019deep,ventura2020can,saunders2021anonysign,CelebAMask-HQ} for sign language rendering \cite{forster2014extensions} could not meet the demands of current sign language rendering due to the increasing volume of training data. Therefore, efficient and high-quality diffusion models are necessary to handle the processing of large-scale American Sign Language (ASL) data.

To accommodate the larger and semantically richer dataset How2Sign, we have developed adaptable modules that optimize mask handling, batch computation, memory utilization, and tensor operations for Fast-SLP Transformers. Additionally, we have designed a novel loss function specifically tailored to the model, which enhances the learning process for longer videos and enables the production of sign language demonstrations of extended durations.
Furthermore, we apply \ourmethodName{} to create higher-resolution and visually appealing representations of real human signers. \ourmethodName{} is the a diffusion model dedicated to producing authentic sign language videos (pose2video). We developed SignDiff by fine-tuning the widely-used Stable Diffusion 1.5 model \cite{Rombach_2022_CVPR}, training it on carefully processed videos from the How2Sign dataset.
%\ourmethodName{} is the first sign language pre-training diffusion model obtained by fine-tuning the standardized SD1.5 \cite{Rombach_2022_CVPR} base model using processed How2Sign video data.
Previous GAN-based sign language rendering methods~\cite{stoll2020text2sign,cui2019deep,ventura2020can,saunders2021anonysign,CelebAMask-HQ} could not meet our demand due to the increasing volume of training data. 
%Therefore, efficient and high-quality diffusion models are necessary to handle the processing of large-scale American Sign Language (ASL) data.

% FR Net
In order to adapt the diffusion model to sign language tasks, we propose a \FrameReinforcementNetLong{} called \FrameReinforcementNet{}. 
Previous sign language rendering approaches, although primarily relying on video frames extracted using the popular human pose estimation algorithm OpenPose \cite{cao2018openpose} as the main input, have also started incorporating DensePose \cite{guler2018densepose} as an auxiliary fine-tuning input (Specifically, DensePose enables precise localization and pose estimation of dynamic individuals by establishing mappings between 2D images and 3D human models). 
%By leveraging deep learning, we map 2D RPG image coordinates to the 3D surface of the human body, segmenting an individual into multiple UV texture coordinates and subsequently processing dense coordinates.
As DensePose \cite{guler2018densepose} is not an inherent conditional input in ControlNet \cite{zhang2023adding}, we devised a DensePose-like input and applied it in \ourmethodName{}. 
The results demonstrate \ourmethodName{} significantly improved image quality, surpassing previous works by more than 10 percentage points.
% Evaluate part
Furthermore, we evaluated our improved text2pose generative model, which exhibited favorable results on both the How2Sign and PHOENIX14T datasets. 
%Our model showcased efficiency and high performance in terms of temporal alignment supervised by \ac{dtw} evaluation and experimental data. 
We also conducted quantitative and qualitative assessments of \ourmethodName{}, and the experiments demonstrated its superior visual quality compared to previous related works \cite{wang2018video,wang2018high,chan2019everybody,saunders2020progressive,stoll2020text2sign,saunders2021mixed}.

\noindent The contributions of this paper can be summarised as:
\begin{itemize}
    \item We propose a larger-scale American sign language pose video generative model, it demonstrates significant improvements in various capabilities, including the generation of long skeletal pose videos.
    \item The novel ASLP approach consists of two new modules and a new loss function, enhances training efficiency, and achieves SOTA performance in text2pose task.
    \item We design a dual-condition \ourmethodName{} model, it introduces a novel DensePose-like novel conditional input by \FrameReinforcementNet{}. This innovation significantly improves pose2video production speed, enhancing image quality, and reducing the number of required training samples.
    %\item The introduction of \FrameReinforcementNet{} into the \ourmethodName{} enables the incorporation of a DensePose-like novel conditional input, resulting in faster pose2video production speed, improved image quality, and the requirement of fewer training samples.
    %\item We conducted comprehensive qualitative and quantitative evaluations to evaluate the proposed approach and models thoroughly.
\end{itemize}

% 这个是介绍文章结构的，但是没地方了，故删掉
%The remaining sections of this paper are organized as follows: Section 2 provides a review of related research on sign language production and label-based image generation. In Section 3, we present a detailed description of our methodology and introduce the FR-Net and SignDiff models, covering their architectures, training procedures, and evaluation methods. Quantitative and qualitative evaluation results for sign language production and image generation are presented in Sections 4 and 5, respectively. Finally, in Sections 6 and 7, we discuss the relevant societal impacts and summarize our contributions.

\section{RELATED WORK}

%\noindent\textbf
\subsection{Sign Language Production}

The development within the field of sign language is not well-balanced, with a concentration of work on \ac{slr} \cite{cui2017recurrent,grobel1997isolated,koller2020quantitative,kadir2004minimal,cooper2007large,koller2015continuous,fang2025signxfoundationmodelsign} and \ac{slt} \cite{camgoz2018neural,camgoz2020sign,ko2019neural,Bohacek_2022_WACV}. 
Various approaches have been explored, including those based on traditional models, neural networks, and model optimization methods. 
These different techniques have been used in the main research directions. 
With the development and deepening of the field, there has been increased attention given to \acf{slp}.

Previously, researchers mainly relied on synthetic animation methods for \acf{slp} \cite{cox2002tessa,karpouzis2007educational,segouat2009study,mcdonald2016automated,10.1145/3490035.3490286,fang2024signllm}. 
However, with the introduction of the Transformer model, many studies have emerged based on neural translation and kinematics. 
These studies typically focused on PHOENIX14T as a means of exploration and experimentation \cite{saunders2020progressive,saunders2021continuous,huang2021towards,saunders2021mixed,saunders2021skeletal}. 
While many of these works involve an intermediate step of translating text to gloss, the temporal control of sign language production itself is challenging, leading to potential inaccuracies when translating to gloss.

In our study, we do not consider this intermediate step due to the significantly larger dataset that our model uses, which does not require the use of an intermediate step to compensate for data scarcity. Additionally, we improve the effectiveness of translation by optimizing structural modules.

\subsection{Rendering of Conditional Input}

Conditioning refers to the ability to control its output according to our intentions like ControlNet \cite{zhang2023adding}. Previous conditional input \acfp{gan} \cite{goodfellow2014generative} showed good performance in image \cite{isola2017image,radford2015unsupervised,wang2018high,zhu2017unpaired,men2020controllable} and video generation \cite{mallya2020world,tulyakov2018mocogan,vondrick2016generating,wang2019few,wang2018video}. There have also been numerous works focusing on generating human poses conditioned on various factors, including whole body \cite{balakrishnan2018synthesizing,ma2017pose,men2020controllable,siarohin2018deformable,tang2020xinggan,zhu2019progressive,chan2019everybody}, face \cite{deng2020disentangled,kowalski2020config,zakharov2019few,10.1007/978-3-030-58517-4_42,8995571,9747380} and hand \cite{liu2019gesture,tang2018gesturegan,wu2020mm}.

One specific application is human-style transfer \cite{NEURIPS2022_ec795aea}, which involves replacing a person in a video with another person while keeping their actions unchanged. This technique has also been widely used in the context of sign language production \cite{chan2019everybody,wei2020gac,zhou2019dance}. The key lies in extracting keypoints to replicate the movements \cite{chan2019everybody,ventura2020can}, with tools such as OpenPose, i3D, and DensePose being commonly employed for keypoint extraction \cite{chan2019everybody,wei2020gac,zhou2019dance,10.1145/3394171.3413532}.

In our work, we propose a novel model called \ourmethodName{}, which incorporates a DensePose-like network named \FrameReinforcementNetLong{} on top of the ControlNet. FR-Net follows the idea of `learning human body shape, not just bone position' and is a good extension based on ControlNet. This model serves as a valuable addition to the field and bridges the gap between diffusion models and sign language.

% \section*{Acknowledgments}

% This document has been adapted
% by Steven Bethard, Ryan Cotterell and Rui Yan
% from the instructions for earlier ACL and NAACL proceedings, including those for
% ACL 2019 by Douwe Kiela and Ivan Vuli\'{c},
% NAACL 2019 by Stephanie Lukin and Alla Roskovskaya,
% ACL 2018 by Shay Cohen, Kevin Gimpel, and Wei Lu,
% NAACL 2018 by Margaret Mitchell and Stephanie Lukin,
% Bib\TeX{} suggestions for (NA)ACL 2017/2018 from Jason Eisner,
% ACL 2017 by Dan Gildea and Min-Yen Kan,
% NAACL 2017 by Margaret Mitchell,
% ACL 2012 by Maggie Li and Michael White,
% ACL 2010 by Jing-Shin Chang and Philipp Koehn,
% ACL 2008 by Johanna D. Moore, Simone Teufel, James Allan, and Sadaoki Furui,
% ACL 2005 by Hwee Tou Ng and Kemal Oflazer,
% ACL 2002 by Eugene Charniak and Dekang Lin,
% and earlier ACL and EACL formats written by several people, including
% John Chen, Henry S. Thompson and Donald Walker.
% Additional elements were taken from the formatting instructions of the \emph{International Joint Conference on Artificial Intelligence} and the \emph{Conference on Computer Vision and Pattern Recognition}.

\begin{figure*}[t!]
    \centering
    \includegraphics[width=1\linewidth]{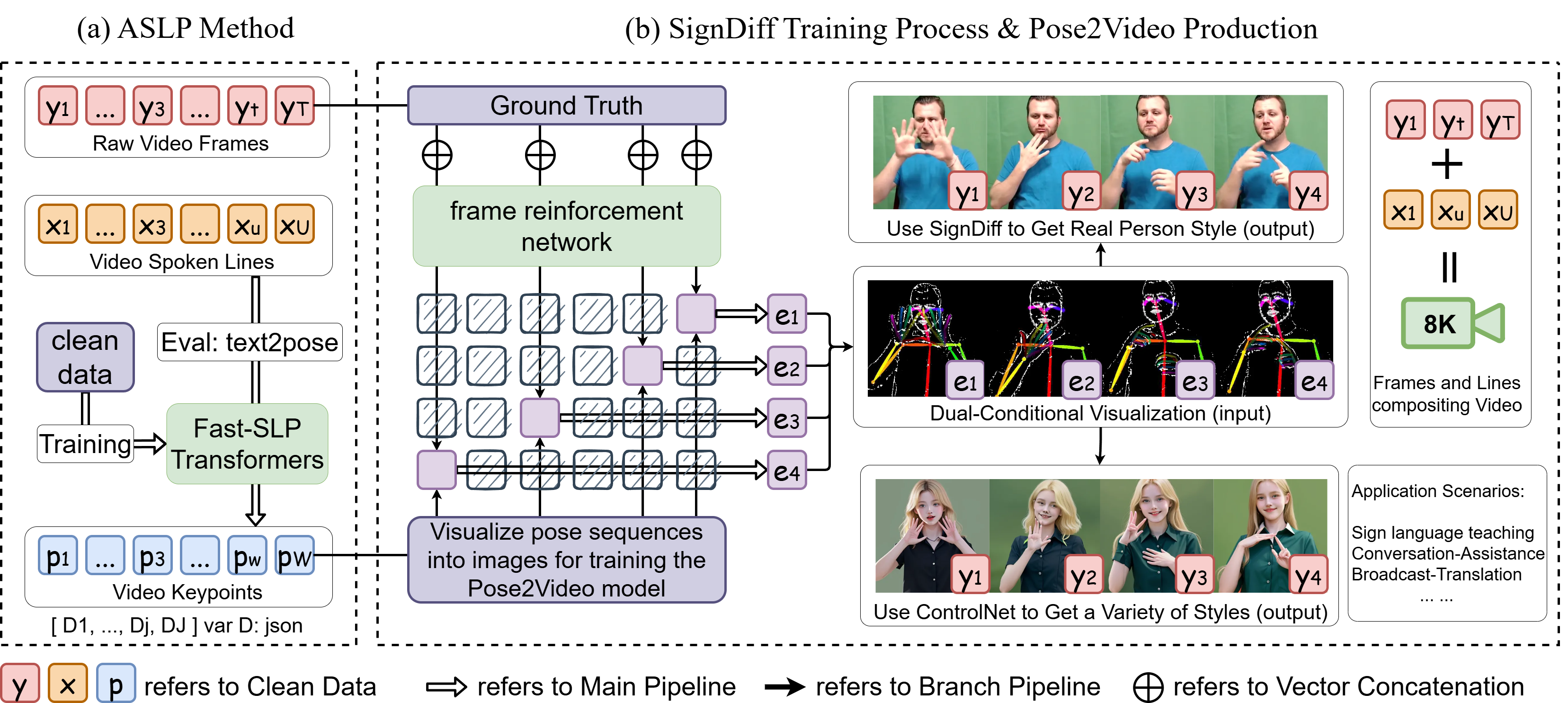}
    \caption{(a) For our \ac{aslp}/text2pose approach, the training data comprises spoken text extracted from videos and pose information ($x$: text, $p$: keypoints json file, $y$: original video frame). (b) The interaction principle of training data and \FrameReinforcementNet{}. When using \ourmethodName{}, the original frames of the video are used to learn human body shape through FR-Net, help \ourmethodName{} realize the function of rendering real human images from visual images of pose key points. $\bm{e}_{1}$ to $\bm{e}_{4}$ represents the output (\ie, dual-conditional intermediate representation, it is the superposition of skeleton and shape). The output of FR-Net can exhibit diverse styles, which are influenced by several factors, including image resolution and fine details.}
    \label{fig:Model_Overview}
    %\vspace{-13pt}
\end{figure*}%

\section{METHODOLOGY}
In this section, we will first give an overview of the Sign Language Production (SLP) pipeline. Then, we will elaborate on the details of our ASLP method and SignDiff.

\subsection{Motivation and Design Idea} \label{sec:Motivation}

For larger-scale \ac{aslp}, the challenge lies in the fact that the previous German gloss vocabulary was around 1000, while our vocabulary exceeds 10000. This necessitates substantial methodological enhancements to accommodate the expanded data. To address this challenge, we have devised and integrated two additional adaptive modules. Furthermore, to effectively handle the heightened semantic complexity, we have introduced a novel loss function tailored for our model.
%To address this, we have built two more adaptive modules and optimized them for masking, batch computation, memory utilization, and tensor calculation. Furthermore, to cope with potential difficulties caused by increased semantic complexity, we propose a new loss adapted to the model. This loss enhances the learning of long videos and makes it feasible to demonstrate long sign language sequences.
Subsequently, GANs were insufficient in simulating genuine and complex sign language data, leading to distorted representations. To overcome this, we adopted diffusion models as advanced sign language rendering models. Unlike GANs, these likelihood-based models effectively capture complex natural image distributions without encountering mode-collapse or training instabilities, all while leveraging parameter sharing instead of excessive parameters like Autoregressive models.

%Lastly, we present the key architecture of ControlNet and conceptually describe how it is trained for fine-tuning and controlling diffusion models. Subsequently, we elaborate on the integration of FR-Net with ControlNet to achieve dual-condition control of diffusion models.

%We approach this problem as a multi-stage sequence-to-sequence task. Firstly, spoken language is translated to sign gloss, \hbox{$\mathcal{Z} = (z_{1},...,z_{\mathcal{W}})$}, as an intermediate representation (Sec. \ref{paragraph:text_to_gloss}). Next, our \FrameSelectionNet{} model co-articulates between gloss-based dictionary signs to produce a full continuous signing sequence, \hbox{$\mathcal{Y} = (y_{1},...,y_{\mathcal{T}})$} (Sec. \ref{sec:gloss_to_sign}). Finally, given $\mathcal{Y}$ and a style image, $\mathcal{S}^{\mathcal{I}}$, our video-to-video signer generation module generates a photo-realistic sign language video, $\mathcal{Z}^{\mathcal{S}}$ (Sec. \ref{sec:sign_to_video}). An overview of our approach can be seen in Fig. \ref{fig:Model_Overview}. In the remainder of this section, we shall describe each component of our approach in detail.

\subsection{Dataset Processing} \label{sec:Dataset_Processing}

The How2Sign dataset is a large-scale multimodal and multiview continuous ASL dataset. It includes a parallel corpus comprising over 80 hours of sign language videos and corresponding modalities such as speech, English transcripts, and depth information. Despite its significance, prior researchers in the field of sign language have not effectively employed this corpus due to the intricacies involved in \acf{aslp}. 
Our objective is to train a Sign Language Pose network to generate sequences of 3D skeleton poses on How2Sign as shown in Fig \ref{fig:Model_Overview}. 
Initially, we extract 2D joint positions from each video using the OpenPose method \cite{cao2018openpose}. 
We then incorporate the skeletal model estimation improvements introduced in Zelink \etal \cite{zelinka2020neural} to transform the 2D joint positions into their corresponding 3D counterparts.
%and run on eight RTX4090 GPUs for ten days. 
Moreover, we employ an iterative inverse kinematics approach to minimize the 3D pose while preserving consistent bone lengths and correcting any misplaced joints. 
Finally, following a similar approach proposed in Stoll \etal \cite{stoll2018sign}, we normalize the skeletal data and adjust the data to a trainable state.
%represent the 3D joints using Cartesian coordinates $x$, $y$, and $z$. 
%Our scripts and preprocessing tools are available in \textbf{supplementary materials} and will be useful to the sign language community.

%Finally, we read the relevant data and delete the unqualified data such as NaN, 0, or replace it with the average median of the data. Finally, we condensed the data to about 1/5 of the original.

%The data required for the diffusion model/Vid2Vid model is processed according to the official documents, and the PHOENIX14T used for the preprocessing is shared by related open sources\footnote{https://github.com/BenSaunders27/ProgressiveTransformersSLP}.
%Our tens of scripts and thousands of lines of preprocessing tools are available in supplementary materials and will be open-sourced.

%We build our Text to Gloss models with 2 layers, 4 heads, and a hidden size of 128 and our \FrameSelectionNet{} with 2 layers, 4 heads, and 64 hidden sizes. We set the interpolation frames, $\mathcal{N}_{LI}$, to 5 and the learning rate to $10^{-3}$. Our code is based on JoeyNMT \cite{JoeyNMT}, and implemented using PyTorch \cite{paszke2017automatic}.

\subsection{Preliminary} \label{sec:production_pipeline}

\noindent\textbf{Text to Pose.} \label{paragraph:text_to_pose}
The general workflow is given a sequence of text, $\mathcal{X}$, our goal is to directly convert it into a continuous signing pose sequence, $\hat{\mathcal{Y}} = (y_{1},...,y_{\mathcal{T}})$ with $\mathcal{T}$ frames. This translation can be formulated as a sequence-to-sequence problem, similar to the text-to-gloss translation discussed in Saunders~\etal \cite{Saunders_2022_CVPR}. However, instead of generating an intermediate gloss sequence, we directly generate the pose sequence.
To achieve this, we utilize and update the encoder-decoder transformer framework used in the gloss-to-pose translation. In the modified framework, the source tokens are the words in the spoken language sequence, and the target tokens are the corresponding pose representations. The encoder-decoder transformer can be defined as follows:

\begin{equation}
    f_{t} = E_{T2P}(x_{t} | x_{1:\mathcal{T}})
\end{equation}
\vspace{-10pt}
\begin{equation}
    p_{w+1} = D_{T2P}(p_{w}  | p_{1:w-1} , f_{1:\mathcal{T}})
\end{equation}

In this formulation, $f_{t}$ and $p_{w}$ represent the encoded source and target tokens respectively, where $p_{w}$ now represents the pose at timestep $w$. Similar to the text-to-gloss translation, we can compute the pose at each timestep as $\hat{y}_{w} = \operatorname*{argmax}_{i} (p_{w})$ until a specific termination condition. 
The $argmax_{i}$ operation is applied over all possible poses, selecting the pose with the highest predicted probability at each time step w. The $argmax_{i}$ denotes the value of i where we maximize the function $p_w$ (the pose probability at time step w).
By directly translating the spoken language sequence to the signing pose sequence, we avoid the need for an intermediate gloss representation. This not only simplifies the translation process but also allows for a more direct mapping between spoken language and sign language. 
%The encoder-decoder transformer framework \cite{vaswani2017attention} can be adapted to accommodate this direct translation, thereby enabling the generation of accurate and articulated signing poses.

%
\begin{figure}[t]
    \centering
    \includegraphics[width=1\linewidth]{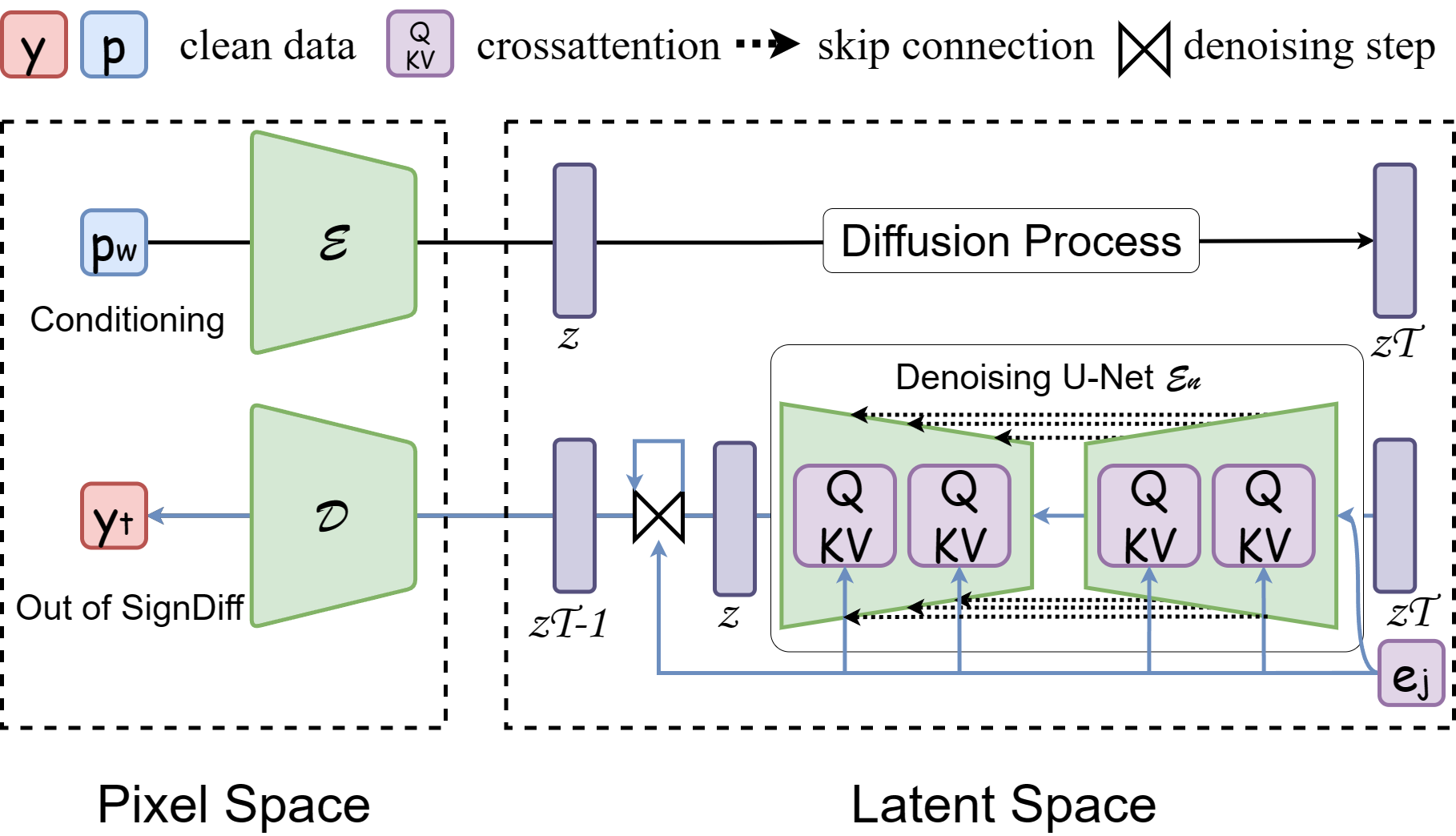}
    \caption{We condition Stable Diffusion (SD) via concatenation or by a more general cross-attention mechanism, which is now the main theoretical basis for controlling SD. $e_j$ is the combined skeleton pose and predicted shape representation, how to get it will be covered in detail in Sec. \ref{sec:SignDiff}.}
    \label{fig:SD}
    \vspace{-12pt}
\end{figure}%

\noindent\textbf{Latent Diffusion Models.} Diffusion Models are probabilistic models that learn a data distribution $p(r)$ by denoising a normally distributed variable in a step-by-step manner \cite{Rombach_2022_CVPR}. This process is the reverse of a fixed $T$ length Markov Chain. Diffusion Models can be seen as a sequence of denoising autoencoders $\model(r_{t},t);\, t=1\dots T$, each trained to predict a clean version of the input $r_t$, as shown in Fig. \ref{fig:SD}. 
\begin{equation}
\lsimple = \expec_{r, \epsilon \sim \mathcal{N}(0, 1),  t }\Big[ \Vert \epsilon - \model(r_{t},t) \Vert_{2}^{2}\Big] \, ,
\label{eq:dmloss}
\end{equation}
The objective of the model is to minimize the difference between the denoised output $r_{t}$ and the original input $x$, with $t$ uniformly sampled from $\{1, \dots, T\}$. This native diffusion model will then be used in ControlNet's base model.

\noindent\textbf{Controllable Pose to Video.}
\label{paragraph:pose_to_video}
ControlNet \cite{zhang2023adding} manipulates the input conditions of neural network blocks to exert control over the overall behavior of the entire neural network. Here, a ``network block" refers to a set of neural layers that are put together as a frequently used unit to build neural networks, \eg, ``resnet'' block, ``conv-bn-relu'' block, multi-head attention block, transformer block, \etc.
Taking the example of a 2D feature, a feature map $\bm{x}\in\mathbb{R}^{h\times w \times c}$ where $\{h, w, c\}$ represent the height, width, and channel dimensions respectively. A neural network block $\mathcal{F}(\cdot;\Theta)$ consisting of parameters $\Theta$ is employed to transform $\bm{x}$ into another feature map denoted as $\bm{y}$. This transformation is represented by the equation $\bm{y}=\mathcal{F}(\bm{x};\Theta)$, and its graphical representation is shown in Figure \ref{fig:Model_Overview}-a and Figure \ref{fig:FR-Net}.

Subsequently, this replica is locked and then cloned, with the clone being trainable. This forms the basis for ControlNet training \cite{ronneberger2015u,isola2017image}, where our \FrameReinforcementNet{} processes the inputs of the cloned replica. The neural network blocks are connected by a unique type of convolution layer called ``zero convolution'', \ie, $1\times 1$ convolution layer with both weight and bias initialized with zeros. 
The resulting model trained through How2Sign is our \ourmethodName{} for pose2video.

\section{IMPLEMENTATION DETAILS} \label{sec:ASL_SLP}

\subsection{Tensor-Optimization-Based New Text2Pose Method} \label{sec:ASL_SLP}

\noindent\textbf{Batch \& Computational Optimization.}
% Batch Optimization
We propose a novel technique, called ``Batch Optimization" to improve the efficiency of mini-batch processing in deep learning architectures. Our approach leverages foundational tensor operations to construct the source mask tensor with appropriate dimensions, reducing the need for extensive tensor manipulations. By directly manipulating the target input tensor, we introduce memory optimization strategies that streamline computations under certain conditions. This avoids the explicit concatenation of tensors within loops and instead utilizes efficient in-place operations. These optimization strategies collectively aim to circumvent memory inefficiency pitfalls, resulting in substantial improvements in memory utilization and computational speed. 
%Empirical evaluations demonstrate the effectiveness of our proposed "batch optimization" technique in significantly advancing the performance of mini-batch processing.

% Computational Optimization
In addition, we introduce a mask optimization technique called ``Computational Optimization" in code to enhance the efficiency of loss computation. The mask, a binary tensor of the same shape as the target tensor, determines whether each corresponding target value is a padding value or not. Applying this mask to the predictions and targets eliminates the need to calculate the loss for padding values, reducing unnecessary computations.
Given a tensor \(T\) and the \(\text{pad\_index}\) used to mark the padding values, we can define a mask tensor \(M\) using the following formula:
\[ M_{ij} = \begin{cases} 
      1, & \text{if } T_{ij} \neq \text{pad\_index} \\
      0, & \text{otherwise}
   \end{cases}
\]
Here, \(i\) and \(j\) represent the indices of the tensor \(T\). The mask tensor \(M\) has the same shape as the target tensor \(T\), and each element in the tensor is 1 if the corresponding element in \(T\) is not a padding value, and 0 otherwise.
%the mask tensor $M$ is defined as $M = (T \neq pad\_index)$, where
Formally, $P$ denotes the tensor of predictions and $T$ denotes the tensor of targets. The masked predictions $P_{masked}$ and masked targets $T_{masked}$ are obtained as $P_{masked} = P \times M$ and $T_{masked} = T \times M$, respectively, using element-wise multiplication ($\times$). The loss is then computed using the selected loss function applied to the masked predictions and targets, such as the default MSE loss or our SA loss.
%In this context, by applying the mask tensor to the prediction tensor and the target tensor, we can eliminate the padding values from the loss computation, thus reducing unnecessary calculations.

%By incorporating these optimizations, we aim to improve the training performance of the model, reduce unnecessary computations, and enhance the convergence rate. These modifications have the potential to accelerate the training process while improving model accuracy and efficiency.

%Given the translated text sequence, $\mathcal{Z}$, we create a stack of ordered dictionary signs, $[\mathcal{D}^{1},...,\mathcal{D}^{w},\mathcal{D}^{\mathcal{W}}]$ (Bottom left of Fig. \ref{fig:Model_Overview}). As in previous works \cite{saunders2021continuous}, we represent each dictionary sign as a sequence of skeleton pose, $\mathcal{D}^{w} = (s^{w}_{1},...,s^{w}_{\mathcal{P}^{w}})$ with $\mathcal{P}^{w}$ frames. We first convert the stack of dictionary signs into a continuous sequence by linearly interpolating between neighboring signs for a predefined fixed $\mathcal{N}_{LI}$ frames. The final interpolated dictionary sequence, $\mathcal{I} = (\mathcal{I}_{1},...,\mathcal{I}_{\mathcal{Q}})$ with $\mathcal{Q}$ frames, is the combination of skeleton pose and the respective linear interpolation.

%\vspace{-12pt}
\noindent\textbf{Scale-Adjusted Loss.}
It aims to improve the generalization of our model and address the overfitting problem when handling extremely large-scale datasets, which is then calculated using a specific criterion, which takes the $preds_{masked}$ and $targets_{masked}$ mentioned in the previous paragraph as inputs.
To adjust the magnitude of the loss, a loss scale factor is applied. If the loss scale is not equal to 1.0, the loss is multiplied by the loss scale factor.
Additionally, $SA_{loss}$ use regularization to prevent overfitting. The regularization loss, $SA_{loss}$, is initialized as a tensor with a value. The $SA_{loss}$ is initialized only before the optimization loop begins. However, $SA_{loss}$ is updated within the loop by adding the SA norm of each weight parameter containing the string 'weight'. The SA norm of the weight parameters in the model is calculated using the torch.norm function. The computed SA norm is then added to the $SA_{loss}$ variable for each weight parameter that contains the `weight' string in its name.
A regularization coefficient, $SA_{lambda}$, is defined with a value of 0.01 to control the strength of the regularization. The SA regularization loss, the result of $SA_{lambda}$ * $SA_{loss}$ will be added to the total loss.

% Scale-Adjusted Loss的废段
%The computed loss is returned as the output of the forward method. This loss function integrates the losses between predicted values and target values along with the regularization loss on the model's parameters. It applies to diverse tasks, including classification, regression, and sequence-to-sequence tasks. By calculating the loss function, we can assess the accuracy and fitting performance of the model on the given dataset, leading to the optimization of model parameters for better overall performance. The effectiveness of our approach relies on the selection of the loss function, as well as the utilization of scaling factors and regularization techniques to adapt to the characteristics of the dataset. Experimental results verify that our approach generally strengthens the understanding and capturing capability of long videos.

\subsection{SignDiff for Efficient Pose2Video Production} \label{sec:SignDiff}

%\vfill\break

\begin{figure}[t]
    \centering
    \includegraphics[width=1\linewidth]{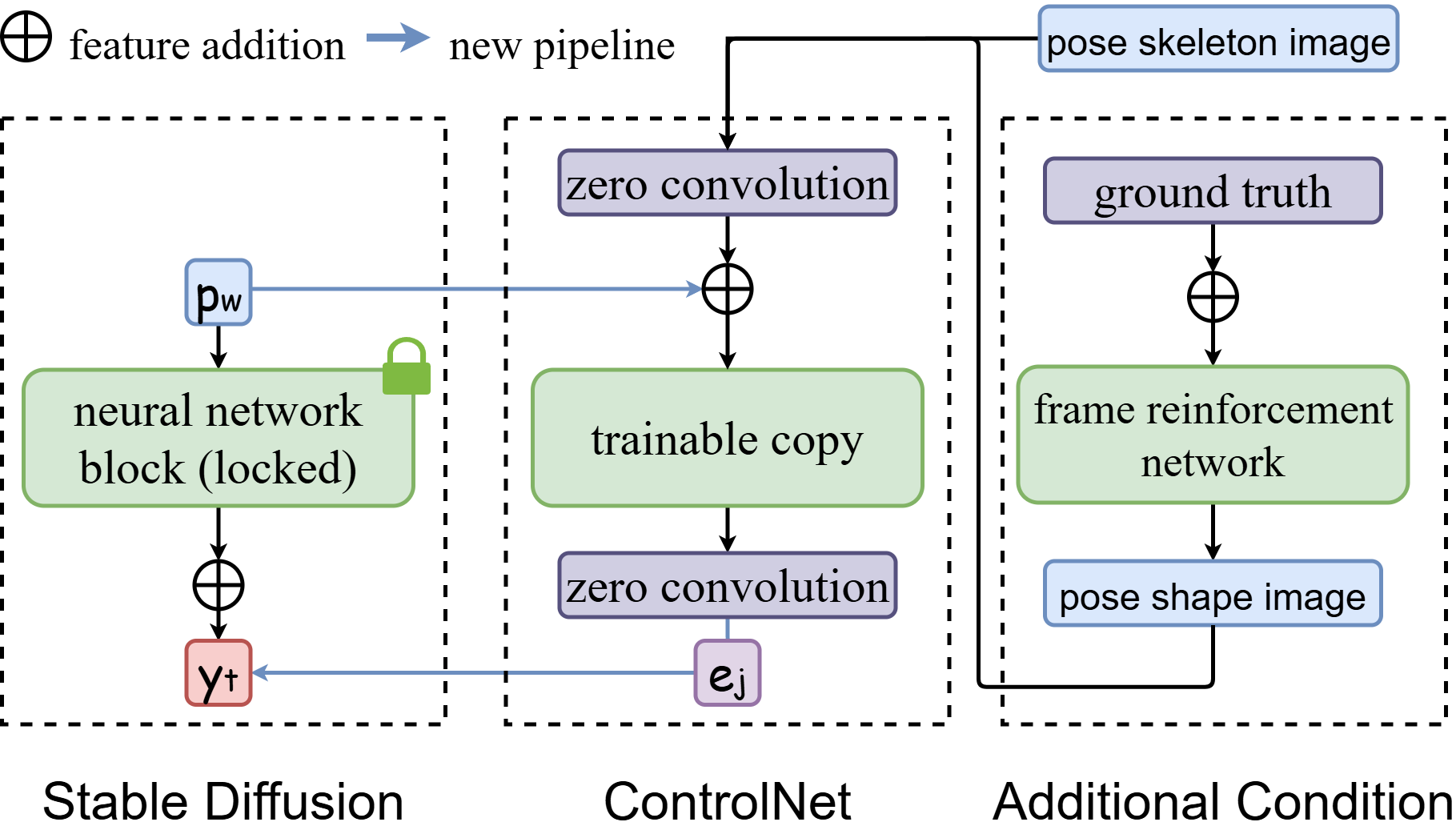}
    \caption{The $\bm{p}_{w}$, $\bm{e}_{j}$ and $\bm{y}_{t}$ represent profound characteristics within neural networks. The term ``zero convolution" denotes a convolution layer with dimensions of 1 × 1, where the weight and bias parameters are initialized to zeroes. This diagram shows how we can add a second extra condition to the ControlNet.}
    \label{fig:FR-Net}
    %\vspace{-12pt}
\end{figure}%

Our SignDiff is used to render visual images of actions generated by ASLP into real images with the same actions. The FR-Net is employed to learn the shape of the human body but not only skeletal pose, creating an intermediate representation during rendering, which enhances the accuracy of synchronizing human body movements.

%\noindent\textbf{Controllable Video Generation.} 
\noindent\textbf{Frame Reinforcement Network}
For SignDiff's training, we modified the ControlNet \cite{zhang2023adding} architecture as shown in Figure \ref{fig:FR-Net}. 
Specifically, we introduced a new conditional input next to ControlNet for capturing the shapes of people. 
The frame data collected from ground truth was processed by \FrameReinforcementNet{} to obtain more specific conditional embeddings. 
This ensures that the generated results are not distorted solely due to the skeletal structure, but also incorporate the learned aspects of human physique. In addition, Controllable SD can reduce the big data crash problems that GANs may encounter.

%\vspace{-8pt}
\FrameReinforcementNet{} utilizes adaptive thresholding to create a binary image based on local pixel thresholds. Subsequently, a 5x5 convolutional kernel is utilized for erosion, shrinking white regions in the binary image. This helps refine foreground object segmentation. ControlNet \cite{zhang2023adding} locks all parameters in $\Theta$ and then clones it into a trainable copy $\Theta_\text{c}$. The copied $\Theta_\text{c}$ is trained with an external condition vector $\bm{c}$.
We extend a copy of $\Theta_\text{d}$ and define the new input as $\bm{d}$.
Here $\bm{c}$ represents OpenPose's input to the model, and $\bm{d}$ represents our FR-Net's input to the model.
Since our input is ultimately superimposed, we can call the variable that $\bm{c}$ and $\bm{d}$ superimpose $\bm{e}$, as shown in Fig. \ref{fig:SD}. Overall, we follow the same principles as ControlNet to incorporate new input conditions.

In principle, as mentioned by the Sec. \ref{paragraph:pose_to_video},
we denote the zero convolution operation as $\mathcal{Z}(\cdot;\cdot)$ and use two instances of parameters $\{\Theta_\text{z1}, \Theta_\text{z2}\}$ to compose the ControlNet structure with
\begin{equation}
	\label{key1}
	\bm{y}_\text{e}=\mathcal{F}(\bm{x};\Theta)+\mathcal{Z}(\mathcal{F}(\bm{x}+\mathcal{Z}(\bm{e};\Theta_\text{z1});\Theta_\text{e});\Theta_\text{z2})
\end{equation}
where $\bm{y}_\text{e}$ becomes the output of this neural network block, as visualized in Fig.~\ref{fig:FR-Net}-middle. The experimental results show that our \FrameReinforcementNet{} can reduce Dynamic Time Warping\footnote{DTW \cite{berndt1994dtw} is the technique used to dynamically warping the time axis of both sequences based on a defined matching criterion.
} scores as well as \methodName{} \cite{Saunders_2022_CVPR}.
%It is achieved by measuring the similarity or distance between the elements at each time point of the sequences. The search for the optimal alignment is performed by dynamically warping the time axis of both sequences based on a defined matching criterion.

%As indicated in Section \ref{paragraph:pose_to_video}, the interconnection between neural network blocks within this framework relies on a distinctive type of convolution layer known as "zero convolution." This specialized layer is implemented through a $1\times 1$ convolution operation, where both the weight and bias are initialized to zero. To represent the zero convolution process, we adopt the notation $\mathcal{Z}(\cdot;\cdot)$. The ControlNet structure encompasses two sets of parameters, denoted as $\{\Theta_\text{z1}, \Theta_\text{z2}\}$, and is composed by combining the outputs of $\mathcal{F}(\bm{x};\Theta)$ and $\mathcal{Z}(\mathcal{F}(\bm{x}+\mathcal{Z}(\bm{c};\Theta_\text{z1});\Theta_\text{e});\Theta_\text{z2})$, as expressed in Equation \eqref{key1}. Consequently, $\bm{y}_\text{e}$ serves as the output of this particular neural network block, depicted in Figure~\ref{fig:FR-Net}-middle.

\begin{table*}[t!]
\centering
\caption{
%Results of The First German Sign Language Production Baseline, and 
Our DL-Based American Sign Language Pose Video Production Baseline.}
\resizebox{0.99\linewidth}{!}{%
\begin{tabular}{@{}p{2.8cm}ccccc|ccccc@{}}
\toprule
 & \multicolumn{5}{c}{DEV SET} & \multicolumn{5}{c}{TEST SET} \\ 
\multicolumn{1}{c|}{Type:} & BLEU-4         & BLEU-3         & BLEU-2         & BLEU-1         & ROUGE          & BLEU-4         & BLEU-3         & BLEU-2         & BLEU-1         & ROUGE          \\ \midrule
%\multicolumn{1}{r|}{Saunders \etal \cite{saunders2020progressive}}     &  11.82 &  14.80 & 19.97 & 31.41 & 33.18 & 10.51 & 13.54 & 19.04 & 31.36 & 32.46 \\
\multicolumn{1}{r|}{\textbf{Fast-SLP (Ours)}}   & {\B 17.19} & {\B 23.11} & {\B 29.49} & {\B 36.96} & {\B 55.85} & {\B 12.85} & {\B 17.35} & {\B 23.38} & {\B 39.46} & {\B 46.89} \\
\bottomrule
\end{tabular}%
}
\label{tab:ourbaseline}
%\vspace{-12pt}
\end{table*}

%\textbf{Ablation Study:} 
\begin{table*}[t!]
\centering
\caption{Fast-SLP Transformer results for Text to Sign Pose production on the How2Sign, with multiple data augmentation techniques. Lines 2 show the techniques from previous work \cite{saunders2020progressive}, and the last two lines are our enhanced methods. FP: Future Prediction, GN: Gaussian Noise, BO: Batch Optimization, CO: Computational Optimization.}
\resizebox{0.99\linewidth}{!}{%
\begin{tabular}{@{}p{2.8cm}ccccc|ccccc@{}}
\toprule

     & \multicolumn{5}{c}{DEV SET}  & \multicolumn{5}{c}{TEST SET} \\ 
\multicolumn{1}{c|}{Approach:}  & BLEU-4         & BLEU-3         & BLEU-2         & BLEU-1         & ROUGE          & BLEU-4         & BLEU-3         & BLEU-2         & BLEU-1         & ROUGE          \\ \midrule

\multicolumn{1}{r|}{Base} & 9.86 & 13.65 & 20.99 & 37.36 & 40.85 & 8.84 & 9.86 & 17.30 & 36.88 & 37.30 \\
%\multicolumn{1}{r|}{Future Prediction} & 12.67 & 17.33 & 25.93 & 43.68 & 45.14 & 12.82 & 15.05 & 23.88 & 41.36 & 43.24 \\
%\multicolumn{1}{r|}{Just Counter} & 14.05 & 20.16 & 27.71 & 45.35 & 47.20 & 12.80 & 15.72 & 25.36 & 43.76 & 44.96 \\
%\multicolumn{1}{r|}{Gaussian Noise} & 15.12 & 21.92 & 29.98 & \textbf{47.32} & \textbf{49.74} & 14.57 & 18.86 & 27.03 & 44.17 & 46.30 \\
\multicolumn{1}{r|}{FP \& GN} & 15.18 & 21.94 & 29.82 & 47.13 & 49.46 & \textbf{15.92} & \textbf{19.65} & \textbf{27.91} & \textbf{46.25} & \textbf{46.57} \\
\multicolumn{1}{r|}{Scale-Adjusted Loss (Ours)}& 14.42 & 20.84 & 28.33 & 44.77 & 47.08 & 15.12 & 18.66 & 26.51 & 44.03 & 44.24\\ 
\multicolumn{1}{r|}{BO \& CO (Ours)}    & \textbf{16.92} & \textbf{22.43} & \textbf{30.01} & 45.15 & 48.75 & 12.28 & 15.99 & 22.67 & 37.58 & 37.84 \\ 
\bottomrule
\end{tabular}%
}
%\vspace{-4pt}
%\caption{\textbf{Ablation Study:} Fast-SLP Transformer results for Text to Sign Pose production on the How2Sign, with multiple data augmentation techniques. FP: Future Prediction, GN: Gaussian Noise, BO: Batch Optimization, CO: Computational Optimization.}
\label{tab:data_augmentation_results}
%\vspace{-12pt}
\end{table*}

\section{EXPERIMENTS} \label{sec:experiments}

In this section, we evaluated our Fast-SLP, \ourmethodName{} and \FrameReinforcementNet{} methods. The experimental content includes back translation assessment, baseline comparison, ablation experiments, and training efficiency tests.

\subsection{Experimental Setup}

As mentioned above, our training is conducted on datasets such as How2Sign and PHOENIX-14T, and we also evaluate our models separately on the same datasets. The How2Sign training dataset consists of 31,048 video segments, with 1,739 segments in the development set and 2,343 segments in the test set. In our preprocessing stage, we first discard text segments that do not have corresponding videos and then batch-separate them into frames. Further processing is applied for different models. The skeletal data is converted into h5 files, followed by filtering out erroneous data (we read the relevant data and delete the unqualified data such as NaN, 0, or replace it with the average median of the data. Then, we condensed the data to about 1/5 of the original). In the end, we obtain 3D skeleton data in the same format as the base model \cite{saunders2020progressive} to train and test our Text2Pose module's methods. Thousands of files are processed and optimized before being read into the final result.

After preprocessing, the How2Sign dataset contains 31,047, 1,739, and 2,343 segments respectively.
%and the PHOENIX-14T dataset contains 7,096, 519, and 642 segments respectively. 
Due to computational limitations and cost considerations, as well as ensuring fair evaluations, we train our models on approximately 8,000 segments from the How2Sign train set. 
%This approach enables comparability across different language families, 
As a simple baseline in ASLP, although it is unnecessary to compare our work with previous ones, we also conducted tests using our method on previous mainstream datasets: 1) For experiments that need to be compared to previous work, we run their PHOENIX-14T dataset, such as the back-translation evaluation in the case of Germanic languages. 2) For experiments that do not need to be compared to previous work, we train and test on How2Sign, such as the ablation experiments, multiple evaluations of \ourmethodName{}, and  Qualitative Evaluation. The training/test data for other models will be mentioned in the Experimental section.

%Our translation model utilizes the video-level translation model developed by Tarrés \etal \cite{slt-how2sign-wicv2023} and translation model \cite{camgoz2020sign}, which are refined with retraining. However, due to their model's lower performance than previous best level of mature sign language interpretation, it fails to fully showcase the accuracy of our modelf in ASL tasks. Thus, we hope that future researchers can further explore and expand in \acf{aslt}.

\subsection{Evaluation for Fast-SLP} \label{sec:ASL_experiments}

\noindent\textbf{Back Translation of ASL.}
Back translation refers to the process of translating generated videos into spoken sentences, which are then compared with the input sentences. 
This serves as a popular metric for evaluating \ac{slp} \cite{saunders2020progressive}. 
BLEU is a commonly used evaluation metric for this method, with higher values indicating a smoother expression of meaning in the videos. 
We conducted tests using our new method on the How2Sign dataset and proposed a baseline for the DL-based ASL video production task, as shown in Table \ref{tab:ourbaseline}.
%(the first row depicts the baseline for the first German Sign Language production task on PHOENIX14 for reference).

%\textbf{Back translation results}
\begin{table}[h]
\centering
\caption{Results on the How2Sign dataset for the \textit{Text to Pose} task, different sign languages have different poses.}
\resizebox{0.99\linewidth}{!}{%
\begin{tabular}{@{}p{2.8cm}cc|cc@{}}
\toprule
     & \multicolumn{2}{c}{DEV SET}  & \multicolumn{2}{c}{TEST SET} \\ 
\multicolumn{1}{c}{Approach:}  & BLEU-4  & ROUGE & BLEU-4 & ROUGE \\ \midrule
\multicolumn{1}{r|}{Progressive Transformers \cite{saunders2020progressive}} & 15.18 & 49.46 & 15.92 & 46.57 \\ 
\multicolumn{1}{r|}{Neural Sign Actors \cite{Baltatzis_2024_CVPR}} & — & — & 13.12 & 47.55 \\ 
\multicolumn{1}{r|}{\textbf{Fast-SLP Transformers (Ours})} & \textbf{17.19}  & \textbf{55.85} & \textbf{12.85} & \textbf{46.89} \\
\bottomrule
\end{tabular}%
}
%\caption{\textbf{Back translation results} on the How2Sign dataset for the \textit{Text to Pose} task, different sign languages have different poses.}
\label{tab:text_to_pose_ASL}
%\vspace{-12pt}
\end{table}

In Table \ref{tab:text_to_pose_ASL}, we also used previous method to train and test it on How2Sign ASL datasets for comparison with previous work. We now use one of these jobs 
%(we didn't try multiple jobs because, as mentioned before, methods are difficult to migrate because sign language data is difficult to process) 
as a baseline to run our new data set for comparison. It shows that our approach is also competitive for excellent latest work, multiple sign languages, not just for data quality reasons. NSA \cite{Baltatzis_2024_CVPR}, which appears after our ASLP work, shows that their performance is only roughly on par with ours (it just 0.27 more).

%\textbf{Ablation study results:}
\begin{figure}[ht]
    \centering
    \includegraphics[width=0.99\linewidth]{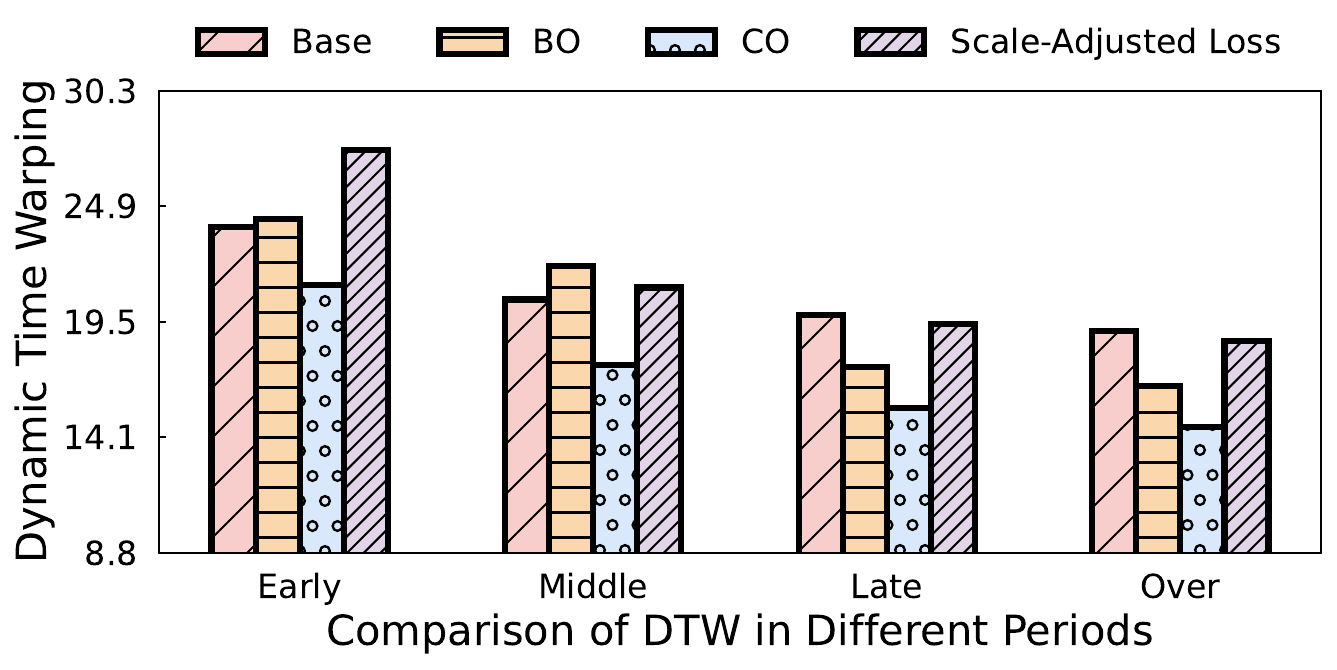}
    %\vspace{-6pt}
    \caption{Comparison of different settings on DTW values (the lower the better) at different training times.}
    \label{fig:former_barline}
    \vspace{-8pt}
\end{figure}%  is determined by epoch

%\vspace{-10pt}
\noindent\textbf{Ablation Study.}
As shown in Table \ref{tab:data_augmentation_results}, firstly, we assess the performance of the model on the text-to-pose task described in Section \ref{paragraph:text_to_pose}. 
We evaluate two new modifications to the baseline model and introduce a new loss function. 
The evaluation is conducted on the development testing set of How2Sign. 
We examine the effectiveness of the baseline model without any modifications and also explore the results of various settings that provide improvements.

In Table \ref{tab:data_augmentation_results}, it can be observed that Gaussian Noise enhances the sensitivity of the model toward isolated word recognition. 
The combination of FP and GN further improves the adaptability of the model to unknown sentences. 
The Scale-Adjusted Loss performs well while emphasizing the temporal characteristics of sign language. This method shows moderate improvements over the base model, particularly in higher-order BLEU scores (BLEU-3 and BLEU-4) on both dev and test sets. Its performance suggests that it effectively captures the sequential nature of sign language, which is crucial for maintaining coherence in longer sign language productions.
Finally, we discovered that our batch module is better suited for longer English sentences, as it achieves a commendable performance.
This is particularly evident in the dev set results, where it outperforms all other methods in BLEU-4 (16.92) scores. These higher-order BLEU scores are more indicative of the model's ability to capture longer phrase structures and overall sentence coherence.
%They were all trained and evaluated on the How2Sign dataset, with other variables such as the learning rate remaining consistent.

\begin{figure*}[t]
    \centering
    \includegraphics[width=0.97\textwidth]{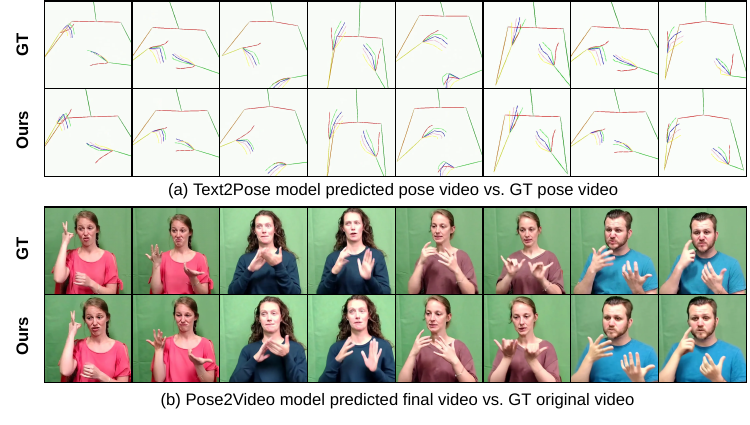}
    \vspace{-6pt}
    \caption{
    %We presented a comparison of \ourmethodName{} with previous work \cite{zhang2023adding,wang2018video}, including (a) input and (b) generated images. We demonstrated the effect of training 10,000 images and 50,000 images, which demonstrates that we only need fewer samples and have higher training efficiency. We also demonstrated the effects that two models of native stable diffusion can achieve for comparison. In reality style, sign language is accurate but prone to missing fingers, while in anime style, fingers are complete but not accurate.
    Our work is divided into two parts: text2pose and pose2video. In this figure, we provide qualitative images from Fast-SLP Transformers and \ourmethodName{} respectively. (a) shows a comparison between predicted pose images and actual pose images, (b) compares our \ourmethodName{} rendered images with the original frames. \ourmethodName{} uses 2D skeleton visualization images as input to ensure fairness, allowing us to focus on comparing our reconstruction quality.
    }
    \label{fig:3group}
    \vspace{-6pt}
\end{figure*}

%\vspace{-8pt}
\noindent\textbf{Training Efficiency Study.}
In Figure \ref{fig:former_barline}, we employ DTW histograms to evaluate the performance of our model during training, without the necessity of complete training. 
The DTW algorithm, based on the concept of dynamic programming, is utilized to calculate the minimum operation distance between sign language videos and sentence sequences. 
It facilitates the computation of visual-textual matching in the latent space, with lower values indicating better results. 
Additionally, we overlay another line plot of the loss during training to comprehensively assess the effectiveness of our modifications and the performance of the model, providing guidance for future work.

%\vfill\break

\begin{table}[h]
\centering
\caption{Back translation results on the PHOENIX-14T dataset for the \textit{Text to Pose} task.}
\resizebox{0.99\linewidth}{!}{%
\begin{tabular}{@{}p{2.8cm}cc|cc@{}}
\toprule
     & \multicolumn{2}{c}{DEV SET}  & \multicolumn{2}{c}{TEST SET} \\ 
\multicolumn{1}{c}{Approach:}  & BLEU-4  & ROUGE & BLEU-4 & ROUGE \\ \midrule
\multicolumn{1}{r|}{Progressive Transformers \cite{saunders2020progressive}} & 11.82 & 33.18 & 10.51 & 32.46 \\ 
\multicolumn{1}{r|}{Adversarial Training \cite{saunders2020adversarial}} & 12.65 & 33.68 & 10.81 & 32.74 \\
\multicolumn{1}{r|}{Mixture Density Networks \cite{saunders2021continuous}} & 11.54 & 33.40 & 11.68 & 33.19 \\ 
\multicolumn{1}{r|}{Mixture of Motion Primitives \cite{saunders2021mixed}} & 14.03 & 37.76 & 13.30 & 36.77 \\
\multicolumn{1}{r|}{Photo-realistic SLP \cite{Saunders_2022_CVPR}}  & 16.92  & 35.74 & 21.10 & 42.57 \\
\multicolumn{1}{r|}{\textbf{Fast-SLP Transformers (Ours})} & \textbf{18.26}  & \textbf{39.62} & \textbf{22.15} & \textbf{46.82} \\
\bottomrule
\end{tabular}%
}
%\caption{\textbf{Back translation results} on the \ac{ph14t} dataset for the \textit{Text to Pose} task.}
\label{tab:text_to_pose}
%\vspace{-12pt}
\end{table}

\noindent\textbf{Back Translation of GSL.} This table \ref{tab:text_to_pose} presents the back translation results for various German Sign Language production methods on the PHOENIX-14T dataset \cite{saunders2020progressive,saunders2020adversarial,saunders2021continuous,saunders2021mixed} for the ``Text to Pose" task. We can observe a clear progression in the field from the earlier Progressive Transformers to our proposed Fast-SLP Transformers. The initial methods such as Progressive Transformers and Adversarial Training achieved relatively lower BLEU-4 and ROUGE scores on both the development and test sets. Subsequently, approaches like Mixture Density Networks and Mixture of Motion Primitives showed improvements in performance. Notably, the Mixture of Motion Primitives method demonstrated significant advancements over its predecessors, they made many useful attempts in accurate pose prediction.

The most recent two approaches, Photo-realistic SLP and our proposed Fast-SLP Transformers, showcase the latest advancements in the field. Photo-realistic SLP achieved a notable performance leap on the test set, particularly in terms of BLEU-4 scores. However, our Fast-SLP Transformers method outperformed all previous work in all metrics, achieving the highest BLEU-4 and ROUGE scores in both development and test sets. The performance of our text2pose model can be directly observed Figure \ref{fig:3group} (a), which indicates that our approach has made substantial progress in generating more accurate and fluent sign language translations, setting a new benchmark for future research in this area.

%Furthermore, Table  showcases the state-of-the-art German Sign Language production methods, where our approach is trained and compared with these works on the  (Because most of them don't have code). The results indicate that even when trained and evaluated on a German dataset, our method achieves competitive performance.

\subsection{Evaluation for SignDiff} \label{sec:SignDiff_experiments}

Finally, we evaluated the image generation performance of \ourmethodName{}, including both qualitative and quantitative evaluations (baseline comparison and ablation and study).

%\vspace{-12pt}
\noindent\textbf{Qualitative Evaluations.} 
%Figure \ref{fig:3group} illustrates three different input types, and final generation results of different models. When comparing our model with Vid2Vid, it is evident that our model achieves better results with less needed data. Upon convergence, our model outperforms the converged state of Vid2Vid in terms of generation quality. Furthermore, when comparing our model with the native diffusion model, we display distinct advantages. While the stable diffusion model exhibits more accurate gestures, it tends to lack finger details. Conversely, the cartoon style, although preserving finger details, demonstrates less precision in gesture representation. Our model combines the strengths of both approaches, offering both accuracy and exceptional image clarity. Additionally, as discussed in Section \ref{sec:Motivation}, our approach exhibits superior image quality upper bounds and enhanced stability.
Figure \ref{fig:3group} (b) shows the quality comparison between our SignDiff reconstructed videos and the original videos. To ensure fairness, we used the same 2D pose visualization images as input, and our reconstruction quality is almost identical to the original. Furthermore, the diffusion model does not produce motion blur, with each frame showing very clear finger details. Our only limitation, as shown in the sixth column, is that hand orientations can become confused, but this minor flaw does not overshadow the overall achievements.

%showcases three distinct input types and the corresponding generation results from different models. A comparison between our model and Vid2Vid reveals that our approach achieves superior results with less data input. Upon convergence, our model surpasses the performance of Vid2Vid in terms of generation quality. Moreover, when contrasted with a common LoRA \cite{hu2022lora} model used to generate human figures, our approach demonstrates clear advantages. While stable diffusion models excel in producing accurate gestures, they often lack detail in finger representation. Conversely, the cartoon style preserves finger details but sacrifices some precision in overall gesture depiction. Our model successfully combines the strengths of both approaches, offering both gesture accuracy and exceptional image clarity. Furthermore, as discussed in Section \ref{sec:Motivation}, our method demonstrates superior image quality upper bounds and enhanced stability. This comprehensive improvement in various aspects of image generation underscores the effectiveness of our approach in addressing the challenges of sign language synthesis.

\noindent\textbf{Baseline Comparison.} We first compared it with previous advanced methods \cite{chan2019everybody,wang2018video,wang2018high,stoll2020text2sign,Saunders_2022_CVPR}, where the input is the corresponding conditional input. 
We compared the output with Ground Truth and calculated the indicator of image quality as follows:
1) Structural Similarity (SSIM) was computed over the entire image \cite{wang2004image}. 2) Hand SSIM was calculated by applying the SSIM metric to a cropped region of each hand. 
3) Hand Pose was determined by measuring the absolute distance between the 2D hand keypoints in the generated and ground truth hand images, utilizing a pre-trained hand pose estimation model \cite{ge20193d}. 4) Fr\'echet Inception Distance (FID) was evaluated for the complete image \cite{heusel2017gans}.

\begin{table}[h]
\centering
%\vspace{-6pt}
\caption{Comparison of Sign Language Video Generation Results: \ourmethodName{} vs. Existing Models.}
\resizebox{0.99\linewidth}{!}{%
\begin{tabular}{@{}p{3.0cm}cccc@{}}
\toprule
  & \multicolumn{1}{c}{SSIM $\uparrow$} & \multicolumn{1}{c}{Hand SSIM $\uparrow$} & \multicolumn{1}{c}{Hand Pose $\downarrow$} & \multicolumn{1}{c}{FID $\downarrow$} \\ \midrule
\multicolumn{1}{r|}{EDN \cite{chan2019everybody}} & 0.737 & 0.553 & 23.09 & 41.54 \\
\multicolumn{1}{r|}{vid2vid \cite{wang2018video}} & 0.750  & 0.570 & 22.51 & 56.17 \\
\multicolumn{1}{r|}{Pix2PixHD \cite{wang2018high}} & 0.737 & 0.553 & 23.06 & 42.57 \\ 
\multicolumn{1}{r|}{Stoll \etal \cite{stoll2020text2sign}} & 0.727 & 0.533 & 23.17 & 64.01 \\ 
\multicolumn{1}{r|}{\methodName{} \cite{Saunders_2022_CVPR} } & 0.759 & 0.605 & 22.05 & 27.75  \\
\multicolumn{1}{r|}{\ourmethodName{} (Ours)} & \textbf{0.849} & \textbf{0.676} & \textbf{20.04} & \textbf{25.22}  \\
\bottomrule
\end{tabular}%
}
%\caption{\textbf{Baseline model comparison} results of \ourmethodName{} sign language video generation.}
\label{tab:baselines_diverse}
%\vspace{-12pt}
\end{table}

%\vspace{-12pt}
%\noindent\textbf{Baseline Comparison.}
%As shown in Table \ref{tab:baselines_diverse}, we compared \ourmethodName{} with the baseline work. We outperformed all previous results, especially in the similarity index between the generated image and Ground Truth, and then combined with Table \ref{tab:ablation_diff}, we can see that it is mainly due to the superiority of the diffusion model itself and our targeted optimization with a large number of high-quality data sets. \FrameReinforcementNet{} plays a very obvious role in this, because it includes the handle shape and not only the skeleton in the training.

Table \ref{tab:baselines_diverse} demonstrates that our \ourmethodName{} significantly outperforms existing state-of-the-art approaches across all evaluation metrics. We achieve substantial improvements in image similarity, with an SSIM score of 0.849 and Hand SSIM of 0.676, surpassing previous benchmarks. Our method also shows superior performance in Hand Pose accuracy (20.04) and perceptual quality (FID score of 25.22).
These improvements can be attributed to the strengths of our diffusion model architecture and targeted optimizations using high-quality datasets. The \FrameReinforcementNet{} plays a crucial role by incorporating detailed hand shape information during training, unlike previous approaches that focus primarily on skeletal structures.
In conjunction with Table \ref{tab:ablation_diff}, these results highlight the effectiveness of our methodological innovations in advancing sign language video generation. Our approach sets a new standard for quality and accuracy in this important field of assistive technology.

\begin{table}[hb]
\centering
%\vspace{-6pt}
\caption{Comparison of our \FrameReinforcementNet{} and other approaches that can enhance image quality \cite{Saunders_2022_CVPR}.}
\resizebox{0.99\linewidth}{!}{%
\begin{tabular}{@{}p{3.0cm}ccccc@{}}
\toprule
  & \multicolumn{1}{c}{SSIM $\uparrow$} & \multicolumn{1}{c}{Hand SSIM $\uparrow$} & \multicolumn{1}{c}{Hand Pose $\downarrow$} & \multicolumn{1}{c}{FID $\downarrow$} \\ \midrule
\multicolumn{1}{r|}{Baseline} & 0.743 & 0.582 & 22.87 & 39.33 \\
\multicolumn{1}{r|}{Hand Discriminator} & 0.738 & 0.565 & 22.81 & 39.22 \\
\multicolumn{1}{r|}{Hand Keypoint Loss} & 0.759 & 0.605 & 22.05 & 27.75 \\
\multicolumn{1}{r|}{\ourmethodName{} (No FR-Net)} & 0.817 & 0.646 & 21.09 & 26.02  \\
\multicolumn{1}{r|}{\FrameReinforcementNet{} (Ours)} & \textbf{0.849} & \textbf{0.676} & \textbf{20.04} & \textbf{25.22}  \\
\bottomrule
\end{tabular}%
}
%\vspace{-4pt}
%\caption{\textbf{Ablation study results:} comparison of our \FrameReinforcementNet{} and others can enhance image quality \cite{Saunders_2022_CVPR}.}
\label{tab:ablation_diff}
% \vspace{}
%\vspace{-12pt}
\end{table}

\begin{figure}[ht]
    \centering
    \includegraphics[width=0.99\linewidth]{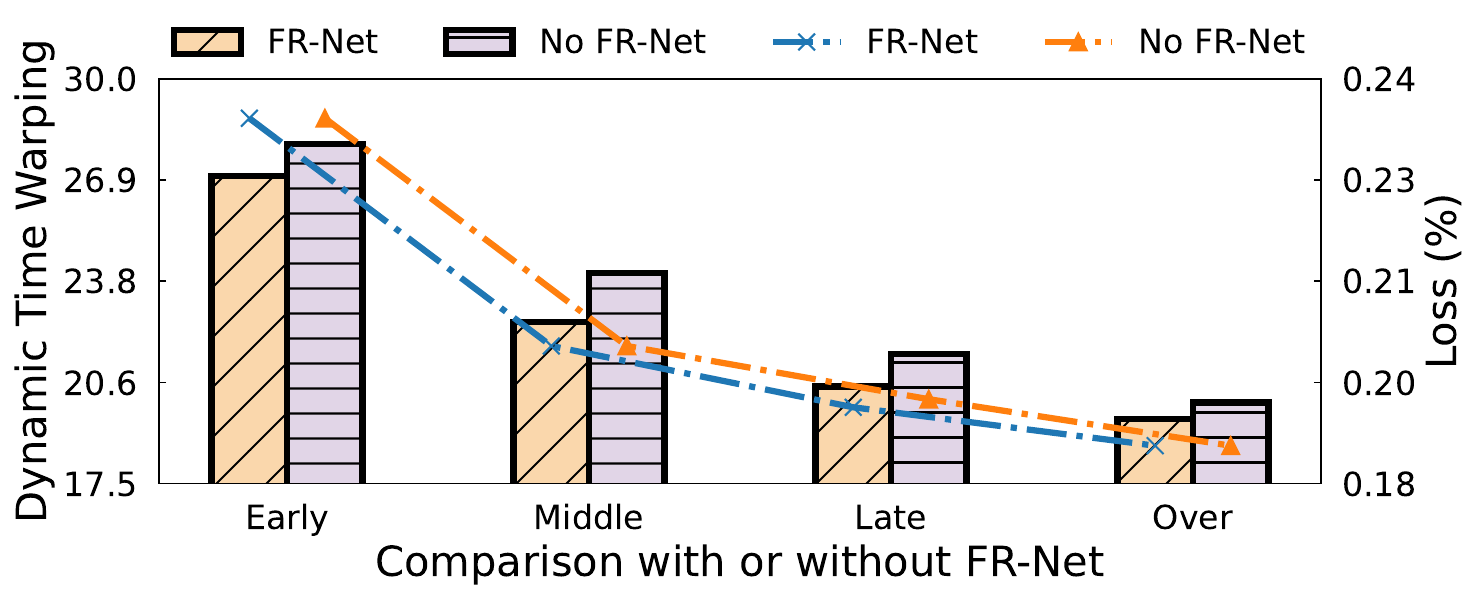}
    %\vspace{-6pt}
    \caption{Comparison of training efficiency between \FrameReinforcementNet{} and \ourmethodName{} (No FR-Net), comparison of the DTW and loss value of different periods (according to the number of pictures trained), the smaller the DTW score, the better.}
    \label{fig:diff_barline}
    %\vspace{-16pt}
\end{figure}%
%\vspace{-12pt}

\noindent\textbf{Ablation Study.}
%We conducted an ablation study on \ourmethodName{} to observe the performance change of our method with or without \FrameReinforcementNet{}, as shown in Table \ref{tab:ablation_diff}. As mentioned by Sec. \ref{sec:intro}, sign language is affected by motion blur in fast-moving images. Our \FrameReinforcementNetLong{} specifically aimed at addressing this issue has indeed improved the overall image quality substantially. 废弃的段落
We conducted an ablation study on \ourmethodName{} to evaluate the impact of various components, particularly the \FrameReinforcementNet{}. Table \ref{tab:ablation_diff} presents a comparison of our method against several baselines and intermediate approaches. The results demonstrate clear performance improvements across all metrics as we incorporate more sophisticated techniques. Our proposed \FrameReinforcementNet{} consistently outperforms other methods, showing substantial enhancements in image quality and hand pose estimation. Compared to the baseline, our method achieves significant improvements: a 10.6\% increase in overall SSIM, a 9.4\% improvement in Hand SSIM, and a 2.83\% reduction in Hand Pose error.
Notably, the FID score shows a dramatic 14.11\% reduction from the baseline to our full model, highlighting the effectiveness of our approach in generating more realistic sign language images. As mentioned in Section \ref{sec:intro}, these improvements address the challenge of motion blur in fast-moving sign language images.

We also assessed the training efficiency of the \FrameReinforcementNet{}, as shown in Figure \ref{fig:diff_barline}. While the rate of decrease in loss remains consistent, we observe a marked improvement in the Dynamic Time Warping (DTW) metric, suggesting enhanced capture of temporal aspects in sign language gestures without significantly increasing computational overhead. In summary, our ablation study validates the effectiveness of the \FrameReinforcementNet{}, demonstrating its crucial role in advancing sign language image enhancement and its potential for improving sign language recognition systems.
%Another evaluative aspect pertains to the training efficiency of the \FrameReinforcementNetLong{}. We conducted an assessment of the training efficiency of the \FrameReinforcementNet{}, examining its impact on model performance and training speed, as depicted in Figure \ref{fig:diff_barline}. It is observed that while there is no significant change in the rate of decrease in loss, the performance metric, DTW, exhibits superior performance. 废弃的段落

\section{DISCUSSION} \label{sec:conclusion}

We note a concurrent work \cite{Baltatzis_2024_CVPR} published before us that adopts an SMPL-based technical approach rather than skeletal poses. Both approaches have their advantages and disadvantages. While the former may have advantages in pure motion generation, for sign language which requires more user-centric approaches, skeletal poses currently can render more realistic and appealing videos for users.

\section{CONCLUSION} \label{sec:conclusion}
Providing high-quality sign language to both learning and user communities is of crucial importance for addressing the shortcomings in the community. 
In this article, we propose a large-scale production model for ASL videos and a pose-to-video model based on diffusion models. 
We introduce a novel \FrameReinforcementNet{} for the deep integration of sign language and diffusion models. 
Additionally, we propose two new modules and a new loss function to adapt to our further developed How2Sign ASL dataset on a larger scale. 
Through baseline comparisons, ablation studies, various parameter analyses, and comprehensive figures, as well as qualitative evaluations, we demonstrate the effectiveness of our approach. In future work, we will enhance the end-to-end capabilities of the model, enabling our technology to be applied to large-scale sign language production.

%%%%%%%%%%%%%%%%%%%%%%%%%%%%%%%%%%%%%%%%%%%%%%%%%%%%%%%%%%%%%%%%%%%%%%%%%%%%%%%%
%\section{ACKNOWLEDGMENTS}

%The authors gratefully acknowledge the contribution of reviewers' comments, etc. (if desired). Put sponsor acknowledgments in the unnumbered footnote on the first page.

%%%%%%%%%%%%%%%%%%%%%%%%%%%%%%%%%%%%%%%%%%%%%%%%%%%%%%%%%%%%%%%%%%%%%%%%%%%%%%%%

%%%%%%%%%%%%%%%%%%%%%%%%%%%%%%%%%%%%%%%%%%%%%%%%%%%%%%%%%%%%%%%%%%%%%%%%%%%%%%%%

\section*{ETHICAL IMPACT STATEMENT}

\textbf{1. Did you read the Ethical Impact Statement Guidelines document?}

Yes, we have read the guidance document on the declaration of ethical influence.

\textbf{2. Is it clear that all studies and procedures described in the paper were approved (or exempted) by a valid ethical review board? Alternatively, is a valid and sufficient justification provided for why the oversight of an ethical review board was not required?}

The data we use is all public data, following the guidelines and licenses provided by the providers. Therefore, for the data itself, we do not assume responsibility, but only as users.

\textbf{3. Does the ethical impact statement provide a clear, complete, and balanced discussion of the potential risks of individual harm and negative societal impacts associated with the research?
%Note that this includes harm to research participants as well as harm to other individuals that may be affected by use, misuse, or misunderstanding of the research.
}

Our \ac{slp} model may contain errors or inaccuracies. Users should exercise caution and verify signs before important use \cite{dickinson2017sign}. We are not responsible for misinterpretations or misunderstandings.

\textbf{4. Does the ethical impact statement describe reasonable, valid, and sufficient use of risk-mitigation strategies by the authors to lessen these potential risks? Alternatively, if relevant strategies were not used, is a valid and sufficient justification for this provided?}

Sign language is nuanced, requiring users to confirm accuracy. We strive to improve our technology but cannot guarantee error-free results. Users must take necessary precautions for clear and accurate communication.

\textbf{5. Does the ethical impact statement provide a valid and sufficient justification for how/why the potential risks of the research are outweighed by the risk-mitigation strategies and potential benefits of the research?
%Note that papers with serious potential risks that are not outweighed by risk-mitigation strategies and potential benefits may be rejected.
}

Our model has strong features tailored for sign language, which will be notably beneficial for the sign language community and researchers. As it stands, it has not reached a fully commercial level yet. Professional signers can quickly discern errors, and novice signers will not rely on it for learning. In a word, continuing research on SLP technology is currently more beneficial than harmful.

\textbf{6a. Does the main paper describe whether/how informed consent and/or assent 
were obtained from participants?}

Not applicable, we follow the license to use public data.

\textbf{6b. Does the main paper state whether the participants explicitly consented to 
the use of their data in the manner described in the paper?}

Not applicable, we follow the license to use public data.

\textbf{6c. Does the main paper explain whether/how participants were compensated?}

Not applicable, we follow the license to use public data.

\textbf{6d. If the research involves any special or vulnerable populations?}

Not applicable, we follow the license to use public data.

%%%%%%%%%%%%%%%%%%%%%%%%%%%%%%%%%%%%%%%%%%%%%%%%%%%%%%%%%%%%%%%%%%%%%%%%%%%%%%%%

%References are important to the reader; therefore, each citation must be complete and correct. If at all possible, references should be commonly available publications.

{\small
\bibliographystyle{ieee}
\bibliography{Ref/main,Ref/sds,Ref/custom}

\begin{thebibliography}{10}\itemsep=-1pt

\bibitem{balakrishnan2018synthesizing}
G.~Balakrishnan, A.~Zhao, A.~V. Dalca, F.~Durand, and J.~Guttag.
\newblock {Synthesizing Images of Humans in Unseen Poses}.
\newblock In {\em Proceedings of the IEEE Conference on Computer Vision and Pattern Recognition (CVPR)}, 2018.

\bibitem{Baltatzis_2024_CVPR}
V.~Baltatzis, R.~A. Potamias, E.~Ververas, G.~Sun, J.~Deng, and S.~Zafeiriou.
\newblock Neural sign actors: A diffusion model for 3d sign language production from text.
\newblock In {\em Proceedings of the IEEE/CVF Conference on Computer Vision and Pattern Recognition (CVPR)}, June 2024.

\bibitem{berndt1994dtw}
D.~J. Berndt and J.~Clifford.
\newblock {Using Dynamic Time Warping to Find Patterns in Time Series}.
\newblock In {\em AAA1-94 Workshop on Knowledge Discovery in Databases}, 1994.

\bibitem{Bohacek_2022_WACV}
M.~Boh\'a\v{c}ek and M.~Hr\'uz.
\newblock Sign pose-based transformer for word-level sign language recognition.
\newblock In {\em Proceedings of the IEEE/CVF Winter Conference on Applications of Computer Vision (WACV) Workshops}, pages 182--191, January 2022.

\bibitem{brentari2018production}
D.~Brentari, J.~Falk, A.~Giannakidou, A.~Herrmann, E.~Volk, and M.~Steinbach.
\newblock {Production and Comprehension of Prosodic Markers in Sign Language Imperatives}.
\newblock {\em Frontiers in Psychology}, 2018.

\bibitem{camgoz2018neural}
N.~C. Camg{\"o}z, S.~Hadfield, O.~Koller, H.~Ney, and R.~Bowden.
\newblock {Neural Sign Language Translation}.
\newblock In {\em Proceedings of the IEEE Conference on Computer Vision and Pattern Recognition (CVPR)}, 2018.

\bibitem{cao2018openpose}
Z.~Cao, G.~Hidalgo, T.~Simon, S.-E. Wei, and Y.~Sheikh.
\newblock {OpenPose: Realtime Multi-Person 2D Pose Estimation using Part Affinity Fields}.
\newblock In {\em Proceedings of the IEEE Conference on Computer Vision and Pattern Recognition (CVPR)}, 2017.

\bibitem{chan2019everybody}
C.~Chan, S.~Ginosar, T.~Zhou, and A.~A. Efros.
\newblock {Everybody Dance Now}.
\newblock In {\em Proceedings of the IEEE International Conference on Computer Vision (CVPR)}, 2019.

\bibitem{camgoz2020sign}
N.~Cihan~Camg{\"o}z, O.~Koller, S.~Hadfield, and R.~Bowden.
\newblock {Sign Language Transformers: Joint End-to-end Sign Language Recognition and Translation}.
\newblock In {\em Proceedings of the IEEE Conference on Computer Vision and Pattern Recognition (CVPR)}, 2020.

\bibitem{cooper2007large}
H.~Cooper and R.~Bowden.
\newblock {Large Lexicon Detection of Sign Language}.
\newblock In {\em International Workshop on Human-Computer Interaction}, 2007.

\bibitem{cox2002tessa}
S.~Cox, M.~Lincoln, J.~Tryggvason, M.~Nakisa, M.~Wells, M.~Tutt, and S.~Abbott.
\newblock {TESSA, a System to Aid Communication with Deaf People}.
\newblock In {\em Proceedings of the ACM International Conference on Assistive Technologies}, 2002.

\bibitem{cui2019deep}
R.~Cui, Z.~Cao, W.~Pan, C.~Zhang, and J.~Wang.
\newblock {Deep Gesture Video Generation With Learning on Regions of Interest}.
\newblock {\em IEEE Transactions on Multimedia}, 2019.

\bibitem{cui2017recurrent}
R.~Cui, H.~Liu, and C.~Zhang.
\newblock {Recurrent Convolutional Neural Networks for Continuous Sign Language Recognition by Staged Optimization}.
\newblock In {\em Proceedings of the IEEE Conference on Computer Vision and Pattern Recognition (CVPR)}, 2017.

\bibitem{deng2020disentangled}
Y.~Deng, J.~Yang, D.~Chen, F.~Wen, and X.~Tong.
\newblock {Disentangled and Controllable Face Image Generation via 3D Imitative-Contrastive Learning}.
\newblock In {\em Proceedings of the IEEE Conference on Computer Vision and Pattern Recognition (CVPR)}, 2020.

\bibitem{dickinson2017sign}
J.~Dickinson.
\newblock {\em {Sign Language Interpreting in the Workplace}}.
\newblock Gallaudet University Press, 2017.

\bibitem{duarte2021how2sign}
A.~Duarte, S.~Palaskar, L.~Ventura, D.~Ghadiyaram, K.~DeHaan, F.~Metze, J.~Torres, and X.~Giro-i Nieto.
\newblock {How2Sign: A Large-Scale Multimodal Dataset for Continuous American Sign Language}.
\newblock In {\em Proceedings of the IEEE/CVF Conference on Computer Vision and Pattern Recognition (CVPR)}, 2021.

\bibitem{fang2025signxfoundationmodelsign}
S.~Fang, C.~Sui, H.~Yi, C.~Neidle, and D.~N. Metaxas.
\newblock Signx: The foundation model for sign recognition, 2025.

\bibitem{fang2024signllm}
S.~Fang, L.~Wang, C.~Zheng, Y.~Tian, and C.~Chen.
\newblock Signllm: Sign languages production large language models, 2024.

\bibitem{forster2012rwth}
J.~Forster, C.~Schmidt, T.~Hoyoux, O.~Koller, U.~Zelle, J.~H. Piater, and H.~Ney.
\newblock {RWTH-PHOENIX-Weather: A Large Vocabulary Sign Language Recognition and Translation Corpus}.
\newblock In {\em Proceedings of the International Conference on Language Resources and Evaluation (LREC)}, 2012.

\bibitem{ge20193d}
L.~Ge, Z.~Ren, Y.~Li, Z.~Xue, Y.~Wang, J.~Cai, and J.~Yuan.
\newblock {3D Hand Shape and Pose Estimation from a Single RGB Image}.
\newblock In {\em Proceedings of the IEEE Conference on Computer Vision and Pattern Recognition (CVPR)}, 2019.

\bibitem{goodfellow2014generative}
I.~Goodfellow, J.~Pouget-Abadie, M.~Mirza, B.~Xu, D.~Warde-Farley, S.~Ozair, A.~Courville, and Y.~Bengio.
\newblock {Generative Adversarial Nets}.
\newblock In {\em Proceedings of the Advances in Neural Information Processing Systems (NIPS)}, 2014.

\bibitem{grobel1997isolated}
K.~Grobel and M.~Assan.
\newblock {Isolated Sign Language Recognition using Hidden Markov Models}.
\newblock In {\em IEEE International Conference on Systems, Man, and Cybernetics}, 1997.

\bibitem{guler2018densepose}
R.~A. G{\"u}ler, N.~Neverova, and I.~Kokkinos.
\newblock Densepose: Dense human pose estimation in the wild.
\newblock In {\em Proceedings of the IEEE Conference on Computer Vision and Pattern Recognition}, pages 7297--7306, 2018.

\bibitem{heusel2017gans}
M.~Heusel, H.~Ramsauer, T.~Unterthiner, B.~Nessler, and S.~Hochreiter.
\newblock {GANs Trained by a Two Time-Scale Update Rule Converge to a Local Nash Equilibrium}.
\newblock In {\em Proceedings of the Advances in Neural Information Processing Systems (NIPS)}, 2017.

\bibitem{huang2021towards}
W.~Huang, W.~Pan, Z.~Zhao, and Q.~Tian.
\newblock {Towards Fast and High-Quality Sign Language Production}.
\newblock In {\em Proceedings of the 29th ACM International Conference on Multimedia}, 2021.

\bibitem{isola2017image}
P.~Isola, J.-Y. Zhu, T.~Zhou, and A.~A. Efros.
\newblock {Image-to-Image Translation with Conditional Adversarial Networks}.
\newblock In {\em Proceedings of the IEEE Conference on Computer Vision and Pattern Recognition (CVPR)}, 2017.

\bibitem{kadir2004minimal}
T.~Kadir, R.~Bowden, E.-J. Ong, and A.~Zisserman.
\newblock {Minimal Training, Large Lexicon, Unconstrained Sign Language Recognition}.
\newblock In {\em Proceedings of the British Machine Vision Conference (BMVC)}, 2004.

\bibitem{karpouzis2007educational}
K.~Karpouzis, G.~Caridakis, S.-E. Fotinea, and E.~Efthimiou.
\newblock {Educational Resources and Implementation of a Greek Sign Language Synthesis Architecture}.
\newblock {\em Computers \& Education (CAEO)}, 2007.

\bibitem{ko2019neural}
S.-K. Ko, C.~J. Kim, H.~Jung, and C.~Cho.
\newblock {Neural Sign Language Translation based on Human Keypoint Estimation}.
\newblock {\em Applied Sciences}, 2019.

\bibitem{koller2020quantitative}
O.~Koller.
\newblock {Quantitative Survey of the State of the Art in Sign Language Recognition}.
\newblock {\em arXiv preprint arXiv:2008.09918}, 2020.

\bibitem{koller2015continuous}
O.~Koller, J.~Forster, and H.~Ney.
\newblock {Continuous Sign Language Recognition: Towards Large Vocabulary Statistical Recognition Systems Handling Multiple Signers}.
\newblock {\em Computer Vision and Image Understanding (CVIU)}, 2015.

\bibitem{kowalski2020config}
M.~Kowalski, S.~J. Garbin, V.~Estellers, T.~Baltru{\v{s}}aitis, M.~Johnson, and J.~Shotton.
\newblock {CONFIG: Controllable Neural Face Image Generation}.
\newblock In {\em Proceedings of the European Conference on Computer Vision (ECCV)}, 2020.

\bibitem{CelebAMask-HQ}
C.-H. Lee, Z.~Liu, L.~Wu, and P.~Luo.
\newblock Maskgan: Towards diverse and interactive facial image manipulation.
\newblock In {\em IEEE Conference on Computer Vision and Pattern Recognition (CVPR)}, 2020.

\bibitem{liu2019gesture}
Y.~Liu, M.~De~Nadai, G.~Zen, N.~Sebe, and B.~Lepri.
\newblock {Gesture-to-Gesture Translation in the Wild via Category-Independent Conditional Maps}.
\newblock In {\em Proceedings of the 27th ACM International Conference on Multimedia}, 2019.

\bibitem{ma2017pose}
L.~Ma, X.~Jia, Q.~Sun, B.~Schiele, T.~Tuytelaars, and L.~Van~Gool.
\newblock {Pose Guided Person Image Generation}.
\newblock In {\em Advances in Neural Information Processing Systems (NIPS)}, 2017.

\bibitem{mallya2020world}
A.~Mallya, T.-C. Wang, K.~Sapra, and M.-Y. Liu.
\newblock {World-Consistent Video-to-Video Synthesis}.
\newblock In {\em Proceedings of the European Conference on Computer Vision (ECCV)}, 2020.

\bibitem{10.1145/3490035.3490286}
S.~Mazumder, R.~Mukhopadhyay, V.~P. Namboodiri, and C.~V. Jawahar.
\newblock Translating sign language videos to talking faces.
\newblock In {\em Proceedings of the Twelfth Indian Conference on Computer Vision, Graphics and Image Processing}, ICVGIP '21, New York, NY, USA, 2021. Association for Computing Machinery.

\bibitem{mcdonald2016automated}
J.~McDonald, R.~Wolfe, J.~Schnepp, J.~Hochgesang, D.~G. Jamrozik, M.~Stumbo, L.~Berke, M.~Bialek, and F.~Thomas.
\newblock {Automated Technique for Real-Time Production of Lifelike Animations of American Sign Language}.
\newblock {\em Universal Access in the Information Society (UAIS)}, 2016.

\bibitem{men2020controllable}
Y.~Men, Y.~Mao, Y.~Jiang, W.-Y. Ma, and Z.~Lian.
\newblock {Controllable Person Image Synthesis with Attribute-Decomposed GAN}.
\newblock In {\em Proceedings of the IEEE Conference on Computer Vision and Pattern Recognition (CVPR)}, 2020.

\bibitem{10.1145/3394171.3413532}
K.~R. Prajwal, R.~Mukhopadhyay, V.~P. Namboodiri, and C.~Jawahar.
\newblock A lip sync expert is all you need for speech to lip generation in the wild.
\newblock In {\em Proceedings of the 28th ACM International Conference on Multimedia}, MM '20, page 484–492, New York, NY, USA, 2020. Association for Computing Machinery.

\bibitem{radford2015unsupervised}
A.~Radford, L.~Metz, and S.~Chintala.
\newblock {Unsupervised Representation Learning with Deep Convolutional Generative Adversarial Networks}.
\newblock {\em arXiv preprint arXiv:1511.06434}, 2015.

\bibitem{Rombach_2022_CVPR}
R.~Rombach, A.~Blattmann, D.~Lorenz, P.~Esser, and B.~Ommer.
\newblock High-resolution image synthesis with latent diffusion models.
\newblock In {\em Proceedings of the IEEE/CVF Conference on Computer Vision and Pattern Recognition (CVPR)}, pages 10684--10695, June 2022.

\bibitem{ronneberger2015u}
O.~Ronneberger, P.~Fischer, and T.~Brox.
\newblock {U-net: Convolutional Networks for Biomedical Image Segmentation}.
\newblock In {\em International Conference on Medical Image Computing and Computer-Assisted Intervention (MIC-CAI))}, 2015.

\bibitem{NEURIPS2022_ec795aea}
C.~Saharia, W.~Chan, S.~Saxena, L.~Li, J.~Whang, E.~L. Denton, K.~Ghasemipour, R.~Gontijo~Lopes, B.~Karagol~Ayan, T.~Salimans, J.~Ho, D.~J. Fleet, and M.~Norouzi.
\newblock Photorealistic text-to-image diffusion models with deep language understanding.
\newblock In S.~Koyejo, S.~Mohamed, A.~Agarwal, D.~Belgrave, K.~Cho, and A.~Oh, editors, {\em Advances in Neural Information Processing Systems}, volume~35, pages 36479--36494. Curran Associates, Inc., 2022.

\bibitem{saunders2020adversarial}
B.~Saunders, N.~C. Camg{\"o}z, and R.~Bowden.
\newblock {Adversarial Training for Multi-Channel Sign Language Production}.
\newblock In {\em Proceedings of the British Machine Vision Conference (BMVC)}, 2020.

\bibitem{saunders2020progressive}
B.~Saunders, N.~C. Camg{\"o}z, and R.~Bowden.
\newblock {Progressive Transformers for End-to-End Sign Language Production}.
\newblock In {\em Proceedings of the European Conference on Computer Vision (ECCV)}, 2020.

\bibitem{saunders2021anonysign}
B.~Saunders, N.~C. Camg{\"o}z, and R.~Bowden.
\newblock {AnonySign: Novel Human Appearance Synthesis for Sign Language Video Anonymisation}.
\newblock {\em arXiv preprint arXiv:2107.10685}, 2021.

\bibitem{saunders2021continuous}
B.~Saunders, N.~C. Camg{\"o}z, and R.~Bowden.
\newblock {Continuous 3D Multi-Channel Sign Language Production via Progressive Transformers and Mixture Density Networks}.
\newblock {\em International Journal of Computer Vision (IJCV)}, 2021.

\bibitem{saunders2021mixed}
B.~Saunders, N.~C. Camg{\"o}z, and R.~Bowden.
\newblock {Mixed SIGNals: Sign Language Production via a Mixture of Motion Primitives}.
\newblock In {\em Proceedings of the International Conference on Computer Vision (ICCV)}, 2021.

\bibitem{saunders2021skeletal}
B.~Saunders, N.~C. Camgoz, and R.~Bowden.
\newblock {Skeletal Graph Self-Attention: Embedding a Skeleton Inductive Bias into Sign Language Production}.
\newblock {\em arXiv preprint arXiv:2112.05277}, 2021.

\bibitem{Saunders_2022_CVPR}
B.~Saunders, N.~C. Camgoz, and R.~Bowden.
\newblock Signing at scale: Learning to co-articulate signs for large-scale photo-realistic sign language production.
\newblock In {\em Proceedings of the IEEE/CVF Conference on Computer Vision and Pattern Recognition (CVPR)}, pages 5141--5151, June 2022.

\bibitem{segouat2009study}
J.~Segouat.
\newblock {A Study of Sign Language Coarticulation}.
\newblock {\em ACM SIGACCESS Accessibility and Computing}, 2009.

\bibitem{siarohin2018deformable}
A.~Siarohin, E.~Sangineto, S.~Lathuiliere, and N.~Sebe.
\newblock {Deformable GANs for Pose-Based Human Image Generation}.
\newblock In {\em Proceedings of the IEEE Conference on Computer Vision and Pattern Recognition (CVPR)}, 2018.

\bibitem{stokoe1980sign}
W.~C. Stokoe.
\newblock {Sign Language Structure}.
\newblock {\em Annual Review of Anthropology}, 1980.

\bibitem{stoll2018sign}
S.~Stoll, N.~C. Camg{\"o}z, S.~Hadfield, and R.~Bowden.
\newblock {Sign Language Production using Neural Machine Translation and Generative Adversarial Networks}.
\newblock In {\em Proceedings of the British Machine Vision Conference (BMVC)}, 2018.

\bibitem{stoll2020text2sign}
S.~Stoll, N.~C. Camg{\"o}z, S.~Hadfield, and R.~Bowden.
\newblock {Text2Sign: Towards Sign Language Production using Neural Machine Translation and Generative Adversarial Networks}.
\newblock {\em International Journal of Computer Vision (IJCV)}, 2020.

\bibitem{7298594}
C.~Szegedy, W.~Liu, Y.~Jia, P.~Sermanet, S.~Reed, D.~Anguelov, D.~Erhan, V.~Vanhoucke, and A.~Rabinovich.
\newblock Going deeper with convolutions.
\newblock In {\em 2015 IEEE Conference on Computer Vision and Pattern Recognition (CVPR)}, pages 1--9, 2015.

\bibitem{tang2020xinggan}
H.~Tang, S.~Bai, L.~Zhang, P.~H. Torr, and N.~Sebe.
\newblock {XingGAN for Person Image Generation}.
\newblock In {\em Proceedings of the European Conference on Computer Vision (ECCV)}, 2020.

\bibitem{tang2018gesturegan}
H.~Tang, W.~Wang, D.~Xu, Y.~Yan, and N.~Sebe.
\newblock {GestureGAN for Hand Gesture-to-Gesture Translation in the wild}.
\newblock In {\em Proceedings of the 26th ACM International Conference on Multimedia}, 2018.

\bibitem{10.1007/978-3-030-58517-4_42}
J.~Thies, M.~Elgharib, A.~Tewari, C.~Theobalt, and M.~Nie{\ss}ner.
\newblock Neural voice puppetry: Audio-driven facial reenactment.
\newblock In A.~Vedaldi, H.~Bischof, T.~Brox, and J.-M. Frahm, editors, {\em Computer Vision -- ECCV 2020}, pages 716--731, Cham, 2020. Springer International Publishing.

\bibitem{tulyakov2018mocogan}
S.~Tulyakov, M.-Y. Liu, X.~Yang, and J.~Kautz.
\newblock {MoCoGAN: Decomposing Motion and Content for Video Generation}.
\newblock In {\em Proceedings of the IEEE Conference on Computer Vision and Pattern Recognition (CVPR)}, 2018.

\bibitem{ventura2020can}
L.~Ventura, A.~Duarte, and X.~Gir{\'o}-i Nieto.
\newblock {Can Everybody Sign Now? Exploring Sign Language Video Generation from 2D Poses}.
\newblock In {\em ECCV Sign Language Recognition, Translation and Production Workshop}, 2020.

\bibitem{vondrick2016generating}
C.~Vondrick, H.~Pirsiavash, and A.~Torralba.
\newblock {Generating Videos with Scene Dynamics}.
\newblock In {\em Advances in Neural Information Processing Systems (NIPS)}, 2016.

\bibitem{wang2019few}
T.-C. Wang, M.-Y. Liu, A.~Tao, G.~Liu, J.~Kautz, and B.~Catanzaro.
\newblock {Few-shot Video-to-Video Synthesis}.
\newblock In {\em Advances in Neural Information Processing Systems (NeurIPS)}, 2019.

\bibitem{wang2018video}
T.-C. Wang, M.-Y. Liu, J.-Y. Zhu, G.~Liu, A.~Tao, J.~Kautz, and B.~Catanzaro.
\newblock {Video-to-Video Synthesis}.
\newblock In {\em Advances in Neural Information Processing Systems (NIPS)}, 2018.

\bibitem{wang2018high}
T.-C. Wang, M.-Y. Liu, J.-Y. Zhu, A.~Tao, J.~Kautz, and B.~Catanzaro.
\newblock {High-Resolution Image Synthesis and Semantic Manipulation with Conditional GANs}.
\newblock In {\em Proceedings of the IEEE Conference on Computer Vision and Pattern Recognition (CVPR)}, 2018.

\bibitem{wang2004image}
Z.~Wang, A.~C. Bovik, H.~R. Sheikh, and E.~P. Simoncelli.
\newblock {Image Quality Assessment: From Error Visibility to Structural Similarity}.
\newblock {\em IEEE Transactions on Image Processing}, 2004.

\bibitem{wei2020gac}
D.~Wei, X.~Xu, H.~Shen, and K.~Huang.
\newblock {GAC-GAN: A General Method for Appearance-Controllable Human Video Motion Transfer}.
\newblock {\em IEEE Transactions on Multimedia}, 2020.

\bibitem{wu2020mm}
Z.~Wu, D.~Hoang, S.-Y. Lin, Y.~Xie, L.~Chen, Y.-Y. Lin, Z.~Wang, and W.~Fan.
\newblock {MM-Hand: 3D-Aware Multi-Modal Guided Hand Generative Network for 3D Hand Pose Synthesis}.
\newblock {\em arXiv preprint arXiv:2010.01158}, 2020.

\bibitem{8995571}
L.~Yu, J.~Yu, M.~Li, and Q.~Ling.
\newblock Multimodal inputs driven talking face generation with spatial–temporal dependency.
\newblock {\em IEEE Transactions on Circuits and Systems for Video Technology}, 31(1):203--216, 2021.

\bibitem{zakharov2019few}
E.~Zakharov, A.~Shysheya, E.~Burkov, and V.~Lempitsky.
\newblock {Few-Shot Adversarial Learning of Realistic Neural Talking Head Models}.
\newblock In {\em Proceedings of the IEEE International Conference on Computer Vision (CVPR)}, 2019.

\bibitem{zelinka2020neural}
J.~Zelinka and J.~Kanis.
\newblock {Neural Sign Language Synthesis: Words Are Our Glosses}.
\newblock In {\em The IEEE Winter Conference on Applications of Computer Vision (WACV)}, 2020.

\bibitem{zhang2023adding}
L.~Zhang and M.~Agrawala.
\newblock Adding conditional control to text-to-image diffusion models, 2023.

\bibitem{9747380}
S.~Zhang, J.~Yuan, M.~Liao, and L.~Zhang.
\newblock Text2video: Text-driven talking-head video synthesis with personalized phoneme - pose dictionary.
\newblock In {\em ICASSP 2022 - 2022 IEEE International Conference on Acoustics, Speech and Signal Processing (ICASSP)}, pages 2659--2663, 2022.

\bibitem{zhou2019dance}
Y.~Zhou, Z.~Wang, C.~Fang, T.~Bui, and T.~Berg.
\newblock {Dance Dance Generation: Motion Transfer for Internet Videos}.
\newblock In {\em Proceedings of the IEEE International Conference on Computer Vision Workshops}, 2019.

\bibitem{zhu2017unpaired}
J.-Y. Zhu, T.~Park, P.~Isola, and A.~A. Efros.
\newblock {Unpaired Image-to-Image Translation using Cycle-Consistent Adversarial Networks}.
\newblock In {\em Proceedings of the IEEE International Conference on Computer Vision (ICCV)}, 2017.

\bibitem{zhu2019progressive}
Z.~Zhu, T.~Huang, B.~Shi, M.~Yu, B.~Wang, and X.~Bai.
\newblock {Progressive Pose Attention Transfer for Person Image Generation}.
\newblock In {\em Proceedings of the IEEE Conference on Computer Vision and Pattern Recognition (CVPR)}, 2019.

\end{thebibliography}
}

\end{document}